\documentclass{article}

\usepackage[nonatbib,preprint]{neurips_2024}

\usepackage[utf8]{inputenc} %
\usepackage[T1]{fontenc}    %
\usepackage{url}            %
\usepackage{booktabs}       %
\usepackage{amsfonts}       %
\usepackage{nicefrac}       %
\usepackage{microtype}      %
\usepackage{xcolor}         %
\usepackage{wrapfig}
\usepackage{enumitem}
\usepackage{graphicx}
\usepackage{float}
\usepackage{amsmath}
\usepackage{amssymb}
\usepackage{array}
\usepackage{multicol,multirow}
\usepackage{setspace}
\usepackage{makecell}
\usepackage{tasks}
\usepackage{pifont}
\usepackage{colortbl}
\usepackage{titletoc}
\usepackage{color}
\usepackage{caption, subcaption, overpic, textpos}
\usepackage{titlesec}

\usepackage[ruled,linesnumbered]{algorithm2e}

\usepackage{algpseudocodex}
\usepackage{xspace}
\newcommand{\modelname}{CLOVER\xspace}
\definecolor{baselinecolor}{gray}{.9}
\definecolor{yellow}{RGB}{218,165,32}
\definecolor{LightSlateBlue}{RGB}{70,130,180}
\definecolor{DeepBlue}{RGB}{65,100,170}
\definecolor{DeepPurple}{RGB}{136,105,160}
\definecolor{LightGreen}{RGB}{59,125,35}
\definecolor{myGreen}{RGB}{77,167,46}
\definecolor{LightRed}{RGB}{227,120,117}
\newcommand{\baseline}[1]{\cellcolor{baselinecolor}{#1}}
\definecolor{robo-orange}{RGB}{255, 191, 0}
\definecolor{deepblue}{RGB}{33, 95, 154}
\definecolor{cvprblue}{rgb}{0.21,0.49,0.74}
\usepackage[pagebackref,breaklinks,colorlinks,citecolor=cvprblue]{hyperref}
\usepackage{cleveref}

\title{\includegraphics[height=0.8em]{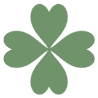} Closed-Loop Visuomotor Control with \\ 
Generative Expectation for Robotic Manipulation}

\author{Qingwen Bu$^{1,2,*}$ \quad Jia Zeng$^{1,*}$\quad
Li Chen$^{1,3,*}$\quad Yanchao Yang$^{3,\natural}$\quad Guyue Zhou$^{4}$\quad \\
\textbf{Junchi Yan}$^{2}$\quad \textbf{Ping Luo}$^{3}$\quad \textbf{Heming Cui}$^{3}$\quad \textbf{Yi Ma}$^{3}$\quad \textbf{Hongyang Li}$^{1,\natural}$ \\
$^{1}$~Shanghai AI Lab 
\quad $^{2}$~Shanghai Jiao Tong University
    \quad
   $^{3}$~HKU\quad $^4$~Tsinghua University\\
  \vspace{-8mm}
}

\begin{document}

\maketitle

\begin{abstract}

Despite significant progress in robotics and embodied AI in recent years, deploying robots for long-horizon tasks remains a great challenge. Majority of prior arts adhere to an open-loop philosophy and lack real-time feedback, leading to error accumulation and undesirable robustness. A handful of approaches have endeavored to establish feedback mechanisms leveraging pixel-level differences or pre-trained visual representations, yet their efficacy and adaptability have been found to be constrained. Inspired by classic closed-loop control systems, we propose CLOVER, a closed-loop visuomotor control framework that incorporates feedback mechanisms to improve adaptive robotic control. CLOVER consists of a text-conditioned video diffusion model for generating visual plans as reference inputs, a measurable embedding space for accurate error quantification, and a feedback-driven controller that refines actions from feedback and initiates replans as needed.
Our framework exhibits notable advancement in real-world robotic tasks and achieves state-of-the-art on CALVIN benchmark, improving by 8\% over previous open-loop counterparts. 
Code and checkpoints are 
maintained at \texttt{\url{https://github.com/OpenDriveLab/CLOVER}}. 
\end{abstract}

\section{Introduction}
\label{sec:intro}
\vspace{-5pt}

Robotics and embodied generalists have gained enormous achievements in recent years, with the successful advancement of representation learning and visual generation in computer vision~\cite{karamcheti2023voltron, zeng2024learning, dasari2023unbiased, wang2023robogen}, large (vision-)language models~\cite{2023embodiedgpt, brohan2023rt2}, policy learning~\cite{reed2022generalist, hafner2019dream}, \textit{etc}. Remarkable behavior intelligence has been demonstrated in diverse and complex single-task settings from picking up a Lego block to solving a Rubik's cube~\cite{haarnoja2018composable, akkaya2019solving}. However, deploying the robots for long-horizon manipulation tasks remains a long-standing challenge~\cite{hu2023toward, li2024foundation}.

\begin{figure}[t]
  \centering
  \includegraphics[width=.95\linewidth]{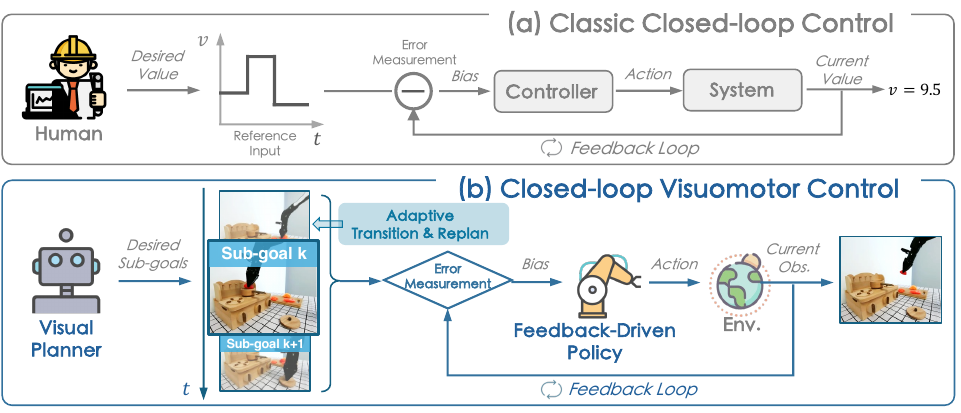}
  \vspace{-3pt}
  \caption{\textbf{Motivation.} The proposed \modelname is inspired by
  the classic closed-loop control in
  automation systems \textbf{(a)}. Our framework \textbf{(b)} employs a visual planner to predetermine a sequence of sub-goals (\Cref{sec:planner}). Then these goals guide the policy to generate actions with an error measurement strategy (\Cref{sec:executor}). Within the feedback loop, it automatically replans when the sub-goal is infeasible, 
  and 
  adapts to
  to the next one upon achievement (\Cref{sec:feedback}).
  }
  \vspace{-0.3cm}
  \label{pipeline}
\end{figure}

Some literature
have attempted to tackle the problem with large language models, splitting a prolonged job into detailed instructions for each minor movement~\cite{driess2023palme, wang2024gensim}, or sub-goals. Though effective in high-level descriptions, texts could still be inadequate for detailed portrayals of the environment and robot state, leading to considerable issues under cross-morphology or multi-environments~\cite{du2024unipi}.
Therefore, 
recent efforts have started embracing vision as a universal medium
to develop an embodied agent capable of planning diverse tasks through imagination and execution~\cite{black202susie, du2024unipi, yang2023unisim}. These approaches involve a generative model for predicting future videos or goal images, followed by a goal-conditioned policy for translating the visual plan into actual actions. 
Despite 
success, they adhere to an \emph{open-loop}
paradigm, 
\textit{i.e.}, proceeding with a fixed sequence of actions without verifying whether the actual trajectory aligns with the planned one. For instance, when a robot is tasked to ``\texttt{grab a Coke from the fridge}'',
current works 
assume that the predicted sub-goal is the visual image of the door opening, and the robot should naturally
achieve the state (sub-goal) prior to grabbing the bottle. 
However, the lack of error measurement and real-time feedback
leads to accumulative deviation, undesirable robustness, and limited adaptability
- 
particularly inadequate for long-horizon tasks and dynamic environments~\cite{hu2023toward, huang2022inner}.
We are inspired by the conventional {closed-loop} control system as depicted in \Cref{pipeline}(a). It aims to regulate physical quantities such as actuator velocity by enhancing control precision via a feedback mechanism. 
Three major components are worth mentioning, the reference input defines desired states which could comprise multiple stages for a prolonged task; the error measurement quantifies bias between the observed state and the planned sub-goal; and the controller adjusts output to reduce the deviation (error)~\cite{hawking1988ACS}. 
In fact, several works have introduced analogous feedback in robotics by measuring errors with pixel appearance~\cite{ebert2018robustness,schenck2017visual} or visual representations~\cite{chen2021learning}, \textit{e.g.}, CLIP features~\cite{radford2021clip}, and yet
their performance
and 
adaptability is limited, leaving accurate error quantification modeling 
unexplored. 

To this end, we propose \textbf{\modelname}, a \textbf{CLO}sed-loop \textbf{V}isuomotor control framework with generative \textbf{E}xpectation for \textbf{R}obotic manipulation. 
As shown in \Cref{pipeline}(b), analogy to 
the classic closed-loop control system, the ingredients in our version are 
adapted
 accordingly.
\textbf{1) Reference inputs.}
With video frames as the interface describing 
desired states, a text-conditioned video diffusion model generates future frames as reference inputs.
To further facilitate the subsequent planning accuracy, we endow the visual planner with the ability of depth map generation, and introduce optical flow regularization to prioritize motion consistency.
\textbf{2) Error measurement.}
Given the limitations of pixel-wise metrics and the inadequacy of pre-trained visual representations, we propose the establishment of a measurable embedding space to realize accurate and efficient error measurement between the observed and planned states. Our state embeddings are trained using an explicit error modeling approach, which yields a strong correlation with the process of converging towards or diverging from target states.
\textbf{3) Feedback-driven controller.}
We present a simple yet effective control framework, comprising a controller and an error-aware adaptive control strategy. The controller is optimized via an inverse dynamics objective~\cite{christiano2016transfer} to achieve the predefined sub-goals.
To address the issue of a goal-oriented policy failing to consistently achieve the desired state, our proposed framework, \modelname, adopts an iterative refinement strategy. 
It continually adjusts its actions to minimize errors, re-evaluates and replans if sub-goals are unrealistic.
To sum up, our {contributions} are three folds: 
\begin{itemize} [leftmargin=*,itemsep=0pt,topsep=0pt]
\item We introduce CLOVER, a generalizable closed-loop visuomotor control framework that incorporates a feedback mechanism to improve adaptive robotic control. %

\item 
We 
investigate the state-measuring attribute for latent embeddings and propose a policy by quantifying feedback errors explicitly. The error quantification settles the construction of an execution pipeline to resolve the challenge of handling uncertainties in video generation and task horizons.

\item Extensive experiments in simulation and real-world robots verify the effectiveness of CLOVER. 
It surpasses prior 
state-of-the-arts
by a notable margin (+8\%) on CALVIN. The average length of completed tasks on real-world long-horizon 
manipulation nearly doubles compared to RT-1.
\end{itemize}

\section{Related Work}
\label{sec:related_work}
\vspace{-5pt}

\noindent\textbf{Closed-loop mechanisms for robotics.} Model-predictive control (MPC) is a classic and popular approach for leveraging learned dynamics for robotic control and has gained great success in learning robust closed-loop policies~\cite{thananjeyan2020safety,erickson2018deep,shin2019autonomous}. Nonetheless, many of these prior works require knowledge about the system state in the planned future which is often infeasible. Visual foresight~\cite{ebert2018robustness,finn2017deep} integrates an auxiliary register network to calculate pixel distances between the current and goal images, consequently providing feedback. And some methods ~\cite{schenck2017visual,kennedy2017precise} utilize an additional detection model with preset rules to estimate the current state. However, the adaptability of these methods is  constrained to single pick-and-pull tasks and does not cover long-horizon, multi-object manipulation tasks. HiP~\cite{ajay2024hip} investigates the feasibility of decomposed language sub-goals and the consistency of generated video plans through training with feedback. Like other visual planners~\cite{black202susie,du2024unipi,yang2023unisim}, it lacks a quantitative assessment of trajectory achievement during test-time execution. 
Inner Monologue~\cite{huang2022inner} leverages vision-language model to provide linguistic feedback for task success detection, but the substantial size of this model hinders the efficiency during test-test execution.
In contrast, CLOVER constructs a measurable latent space from pixel observations and qualitatively measures deviations from planned goals for each action, thereby incorporating real-time feedback. 
Furthermore, the feedback mechanism is incorporated into long-horizon manipulation tasks.
\noindent\textbf{Diffusion model as a visual planner.} 
Recently, it is trending 
to utilize diffusion models as visual planners to generate goal states. UniPi~\cite{du2024unipi} seminally leverages internet data to train a text-conditioned video generator and uses an inverse dynamics model to estimate ultimate actions. UniSim~\cite{yang2023unisim} creates a universal video diffusion model for simulating interactions and training policies through generative modeling. Ajay \textit{et al.}~\cite{ajay2024hip} propose compositional foundation models for hierarchical planning, including task decomposition, visual planning, and action inference. They also utilize an additional 
classifier
to deal with the uncertainty of generation quality.
SuSIE~\cite{black202susie} utilizes an image-editing model as a high-level planner to set achievable sub-goals for a low-level controller, 
while
ADVC~\cite{Ko2023Learning} infers actions from predicted video content with dense correspondences.
Though prior techniques can synthesize visually reasonable future sub-goals, one challenge is that the lack of consistency constraints related to geometry and motion potentially diminishes the fidelity of generated videos for policy prediction and increases generation instability.
In our work, an RGB-D video prediction model is introduced and constrained by the optical flow to enhance sub-goals' reliability. Moreover, the instability of visual plans from diffusion models is rarely discussed in the aforementioned literature. Contrarily, at the core of \modelname, we adopt an
policy state estimation to 
detect unreachable plans by measuring the distance between consecutive frames.

\section{Methodology}
\label{sec:method}
\vspace{-5pt}

We aim at building a generalizable framework that integrates the closed-loop philosophy into robotic visuomotor control. The overall system is illustrated in \Cref{pipeline}. Accordingly, in order to set the desired value before execution, we introduce a visual planner that generates consecutive sub-goals (\Cref{sec:planner}). In \Cref{sec:executor}, we detail the structure of our feedback-driven policy to decode actions, and demonstrate how to measure the deviation from the current to the goal states. Finally, the overall test-time execution pipeline of \modelname is leveraged in \Cref{sec:feedback}.

\subsection{Visual Planner}
\label{sec:planner}
\vspace{-5pt}

The visual planner is to produce a reliable sequence of future plans based on the initial observation $O_{0}$ and task descriptions $c_{l}$. Inspired by previous successful attempts~\cite{du2024unipi,Ko2023Learning,yang2023unisim}, we also employ the prevailing conditional diffusion model to realize text-conditioned video generation~\cite{singer2022make}. Derived from the image diffusion model of Imagen~\cite{saharia2022imagen,Ko2023Learning}, our model is designed to generate future videos (\textit{i.e.}, predicted sub-goals) spanning predetermined time frames, denoted as $\{ \hat{O}_{1},\hat{O}_{2},...,\hat{O}_{K}\}$, with $K=8$.
However, different from targets of high-resolution and meticulous structures for general generative models, visual planners for robotic manipulations highlight the need to understand spatial environments and robot movements. Therefore, designs of effectively integrating depth information and leveraging optical flow's regularization are introduced in \modelname to generate geometry-aware and temporally coherent futures, which we describe below.

\noindent\textbf{Text-conditioned RGB-D video generation.} 
In the framework of amalgamating video prediction and goal-conditioned policy modules, generating a visual plan that precisely corresponds to the task description is a prerequisite for accomplishing manipulation tasks. To encode language inputs, we employ the tokenizer and encoder from CLIP~\cite{radford2021clip} as the basis, following~\cite{ramesh2022hierarchical}.
In addition to the condition injection techniques outlined in Imagen~\cite{saharia2022imagen}, which integrate language embeddings into the latent space of the diffusion model directly, our model further incorporates cross-attention-based conditioning to enhance its language-following ability. Moreover, utilizing classifier-free guidance~\cite{ho2022classifier}, the visual planner demonstrates encouraging controllability and generalization, being able to produce diverse and reasonable plans based on task descriptions (See analysis in \Cref{sec:exp-results}).

For the vision inputs, robots operating in the 3D space face great challenges in learning from 2D observations directly~\cite{kent2020leveraging}. Therefore, considering the ease of acquisition of depth sensory nowadays in robotics and its accurate spatial depiction of the environment, we incorporate geometric information from depth maps to assist in manipulation.
To predict RGB-D videos, we adopt a simple yet effective way. Specifically, the RGB image and depth map are concatenated on the channel dimension and embedded into a unified latent space throughout all layers of the model. Compared to devising distinct branches for each modality, this yields satisfactory generation results with high consistency between modalities in practice. Moreover, the straightforward approach opens the potential for pre-training the diffusion model on large-scale RGB-only datasets to further enhance its capabilities~\cite{wang2022pretraining, zhang2023adding}. 

\noindent\textbf{Latent regularization with optical flow.} Besides the easy acquisition of the depth modality, the robotic manipulation tasks also feature in their interaction dynamics, \textit{i.e.}, the moving robot arm and interacted objects in the environment.
Though existing works~\cite{black202susie,ajay2024hip} have utilized video diffusion models for visual plan generation, they fall short in considering the essential gaps between robot manipulation and general video data adequately, particularly the static camera position and robot-initiated movements~\cite{padalkar2023open}. 
Drawing inspiration from the importance of motion cues in robot manipulation, we propose to incorporate optical flow as an explicit regularization term to further foster the classic video diffusion models for manipulation tasks. Specifically, following the end-to-end optical flow estimation framework, RAFT~\cite{teed2020raft}, we first build the pixel-wise correspondence map between the diffusion latent of two consecutive frames. This map is then utilized by subsequent modules to iteratively refine the optical flow estimation through lookup and update operations. Given the final estimation, our flow-based regularization term is formulated as:
\begin{equation}
    L_{\text{reg}} = \frac{1}{K-1} \sum_{k=1}^{K - 1} \Vert O_{k+1} -  \mathcal{W}(O_{k}, \hat{F}_{k \rightarrow k+1})\Vert ,
\end{equation}
where $\hat{F}_{k \rightarrow k+1}$ is the estimated optical flow, $\mathcal{W}$ represents the wrapping function and $\{O_{k}, O_{k+1}\}$ are two consecutive frames in the ground-truth video. More details are provided in~\Cref{appendix:impl_details}.

\begin{figure}[t!]
    \centering
    \includegraphics[width=0.95\linewidth]{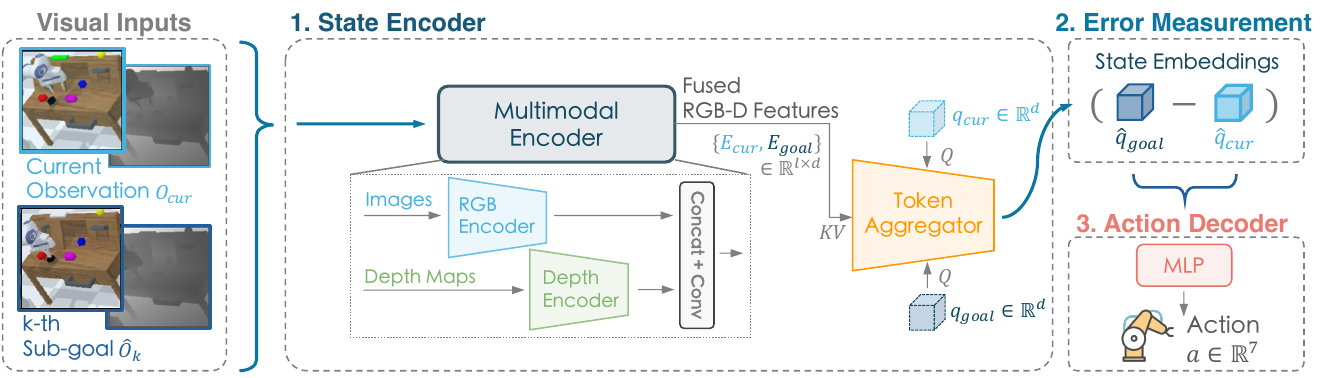}
    \vspace{-3pt}
    \caption{\textbf{Architecture of our feedback-driven policy.} \textbf{1)} The state encoder takes in both current observation along with the synthesized sub-goal. A shared multimodal encoder generates fused RGB-D features, followed by two queries extracting informative features as the current and goal embeddings respectively. \textbf{2)} The discrepancy of the two state embeddings is explicitly modeled as errors.  \textbf{3)} The resultant residual in error measurement is ultimately decoded to the final action.}
    \label{fig:policy}
    \vspace{-5pt}
\end{figure}

\subsection{Feedback-Driven Policy}
\label{sec:executor}
\vspace{-5pt}

As depicted in \Cref{fig:policy}, from current and desired visual inputs to the ultimate action output, our policy can be divided into the following components: 1) State Encoding: Deriving informative features from visual inputs and producing compact state embeddings that encode current and desired sub-goal states; 2) Error Measurement: Formulating the deviation from current to goal state; 3) Action Decoding: Decoding the deviation signal into the action a robot can actuate. 

\noindent\textbf{State encoder.} To begin with, we employ a multimodal encoder to transform raw pixel inputs into enriched visual representations, which comprises two ViT-based~\cite{dosovitskiy2020image} encoders for RGB and depth respectively along with a multimodal feature fusion module. The feature fusion process uses squeeze-and-excitation module~\cite{hu2018squeeze} for channel-wise integration and selection. Subsequently, the token aggregator adaptively selects critical information pertaining to manipulation from the sequence of visual features, condensing them into a compact state embedding. Expressly, the token aggregator is built upon a multi-head attention pooling~\cite{lee2019set} with a two-layer multi-layer perceptron (MLP) performing nonlinear projection. Given fused RGB-D features $\{E_{\text{cur}},\ E_{\text{goal}}\} \in \mathbb{R}^{l\times d}$ corresponding to current and goal inputs, respectively, this process can be specified as:
\begin{equation}
    \hat{q} = \texttt{MLP} \big[ \texttt{MultiHeadAttn}(Q = q, K = V = E  ) \big ], \\
\end{equation}
where $\hat{q} \in \{\hat{q}_{\text{cur}}, \hat{q}_{\text{goal}} \}$ denotes state embeddings and $q \in \{q_{\text{cur}}, q_{\text{goal}}\}$ are queries initialized to extract visual features $E \in \{E_{\text{cur}},\ E_{\text{goal}}\}$. Here $d$ is the hidden dimension size and $l$ represents the visual token length. We employ shared weights for encoding both states in parallel, with the exception that two separately initialized queries are utilized to extract each state embedding. State encoder gives rise to an information bottleneck, prioritizing the encoding of manipulation-relevant features while filtering out irrelevant background details to ensure informativeness.

\begin{figure}[t]
    \centering
    \includegraphics[width=.96\linewidth]{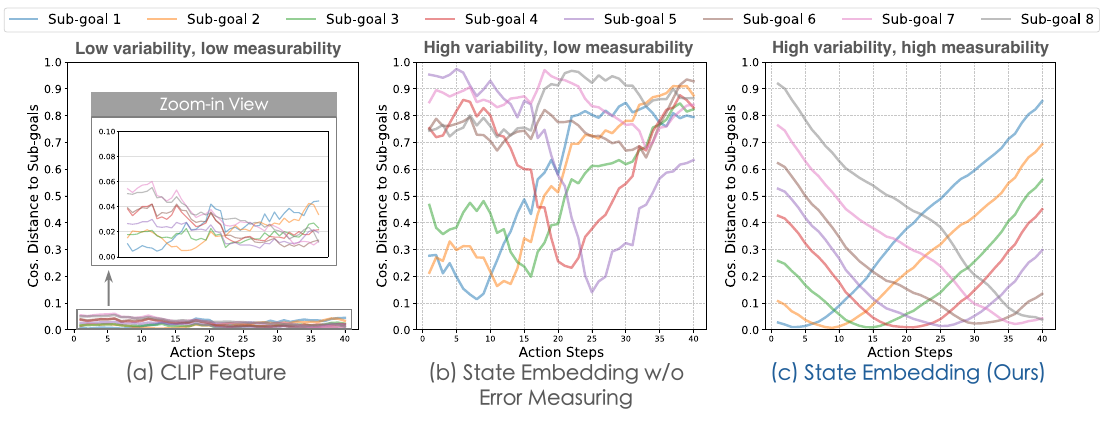}
    \vspace{-10pt}
    \caption{\textbf{Comparison on the measurement ability of
    different embeddings.}
    We visualize the cosine distance between embeddings of observations and generated sub-goals during a roll-out process. %
    \textbf{\color{gray}(a)} CLIP feature~\cite{radford2021clip} and \textbf{\color{gray}(b)} state embeddings trained without error measuring do not hold clear interrelations among frames. While \textbf{\color{deepblue}(c)} state embeddings obtained from our policy distribute reasonably in the latent space which benefits measuring the errors in feedback loops.}
    \label{fig:explain_closed_loop}
    \vspace{-8pt}
\end{figure}

\noindent\textbf{Error measurement.} In conventional closed-loop control systems, the controller generates control signals based on the error between the desired value and feedback signals~\cite{cheon2015replacing}. Analogously, in our visuomotor control pipeline, we explicitly model the discrepancy between the current and goal states by performing element-wise subtraction of the two corresponding state embeddings. Despite its simplicity, this approach has been proven effective in practice. It induces a strong prior that latent actions formulate the transitions between latent states~\cite{hafner2019dream}.
More importantly, the transitions can be quantified and we note that not all embeddings hold the essential characteristics.

Notably, learning representations to distinguish among diverse instructions and visual states has been a long-standing topic~\cite{pmlr-v70-pathak17a}, while there exist few works exploring their quantitative metric for action. In \modelname, the measurement capability of state embeddings is observed by learning to act from deviation signals, which is absent in pre-trained visual encoders or policy models learned based on current observations solely, \textit{i.e.}, behavior cloning. As illustrated in \Cref{fig:explain_closed_loop}(c), the cosine distance between state embeddings decreases together with the robot approaching each predicted sub-goal. In the meantime, the distance to the previous sub-goal increases as proceeding to the next one. Besides, the numerical range of the distance spans approximately from 0 to 0.9, thereby providing a sufficient margin for distinguishing and identifying current states. On the contrary, concerning visual representations generated by pre-trained encoders~(\textit{i.e.}, CLIP features in \Cref{fig:explain_closed_loop}(a)), there is a noticeable reduction in the range of value variations (within 6e-2) with pronounced fluctuations, although the curves show similar patterns in general. This result stands for all the pre-trained visual encoders we have studied, owing to the fact that manipulation-relevant features will be overwhelmed by immaterial background information. Furthermore, we also test employing state embeddings generated by our policy model but optimized without error measuring. The resulting embeddings capture the state propagation as evidenced by the significant numerical variability (\Cref{fig:explain_closed_loop}(b)); however, they lack the capability to measure interrelations among sub-goals and exhibit poor monotonicity when reaching each sub-goal. Next, we introduce how to elevate the satisfactory error measuring feature for action decoding and adaptive feedback control autonomously.

\noindent\textbf{Action decoder.} To keep the framework concise and generalizable, we simply adopt an MLP to decode action outputs from error signals. We consider the action space of a 7-DoF robotic arm, encompassing the position of the end-effector $a_{\text{EE}} \in \mathbb{R}^6$ and the gripper state $a_{\text{griper}} \in \mathbb{R}^1$. Our policy is learned with an Inverse Dynamics objective $\pi_{\phi}(a_{0} | O_{0}, O_{k})$, where it infers action $a_{0}$ based on current observation $O_{0}$ and specified sub-goals $O_{k}$. To promote the transferability of our framework, we exploit third-view RGB-D images as inputs only, with action labels as the training targets. State information such as proprioception signals or gripper-view images are not applied to facilitate manipulation. Please refer to \Cref{appendix:impl_details} for further architectural and training details.

\begin{center}
\vspace{-6pt}
\begin{minipage}[t!]{\linewidth}
\footnotesize
\begin{algorithm}[H]
  \SetKwInOut{Parameter}{Hyper parameters}
  \SetAlgoLined
  \KwIn{Visual planner $p_{\theta}$; Policy $\pi_{\phi}$; State encoding module $g_{\phi}(\cdot)$; Cosine distance $D_C(\cdot, \cdot)$.}
  \Parameter{Time limit $T$; Distance threshold for replan and sub-goal transition $\{D_{R}, D_{S}\}$.}
  \BlankLine

   $t \leftarrow 0, i_{\text{sub}} \leftarrow 0$ \hfill $ \rhd\ \text{\small \color{gray}Initialize the sub-goal selection index}$\\
  \While{$t \leq T$}{
  \If{\text{Replan} \textbf{or} $t == 0$}{
    $\hat{O}_{1:K} \sim p_{\theta}(O_{1:K}\ | \ O_{0}, c_{l})$  \hfill $\quad \rhd\  \text{\small \color{gray}Generate language-conditioned sub-goals (\Cref{sec:planner})}$ \\
    \eIf{$ \max\limits_{ k = 1,...,K-1}\{D_C(g_{\phi}(\hat{O}_{k}),\ g_{\phi}(\hat{O}_{k+1})) \} > D_{R}$}{ \textit{Replan} $\leftarrow$ \textit{True} \hfill  $\rhd \text{\small \color{gray}Replan if sub-goals are unreachable}$} 
    {\textit{Replan} $\leftarrow$ \textit{False}} 
}
  \If{$D_C(g_{\phi}(O_{0}),\ g_{\phi}(\hat{O}_{i_{\text{sub}}})) < D_{S}$}{
  $ i_{\text{sub}} \leftarrow i_{\text{sub}} + 1$  \hfill $ \quad \rhd\  \text{\small \color{gray}Transition if the current sub-goal has been reached} $
  }
  $\text{Sample and Execute}\  \hat{a} \sim \pi_{\phi}(a_{0} | O_{0}, \hat{O}_{i_{\text{sub}}}) \hfill  \rhd \text{\small \color{gray}Predict and execute action (\Cref{sec:executor})}$ \\
  $O_{0} \leftarrow \textit{Env}(\hat{a})$ \hfill $ \quad \rhd\  \text{\small \color{gray}Update current observation} $\\
  $t \leftarrow t + 1$
 }
  \caption{CLOVER: Test-time Execution}
  \label{alg:execution}
\end{algorithm}
\end{minipage}
\end{center}

\subsection{\modelname}
\label{sec:feedback}
Equipped with the error quantification capability aforementioned, we have developed a closed-loop visuomotor control framework with feedback, illustrated in \Cref{alg:execution}. Notably, our framework distinguishes itself through two key aspects: 1) It can detect and address the instability of diffusion models by initiating replanning when sub-goals are unreachable; 2) It achieves adaptive transitioning between sub-goals based on the distance measurement. 
In contrast to previous literature such as SuSIE~\cite{black202susie} that sets up dataset-dependent hyperparameters to manually regulate sub-goal refreshing (usually required to be consistent with the frame intervals during training) and thus potentially limit their performance and scalability, our proposed \modelname is agnostic to training details and adaptable to visual planners with varying intra-frame intervals. We provide illustrative examples in \Cref{appendix:illustration} to demonstrate the functionality of replanning and adaptive sub-goal transitions. 

\section{Experiments}
\label{sec:exp}
\vspace{-5pt}

\begin{table}[t]
    \centering
    \caption{\textbf{Long-horizon visuomotor control on CALVIN ABC$\rightarrow$D.} We report success rates along with the average length of completed tasks (out of the whole 5 tasks) per evaluation sequence. CLOVER outperforms all previous methods by a notable margin.
    \textit{Lang} and \textit{All} denote whether models are trained only with the subset vision-language data.  $^{*}$Results reported by~\cite{black202susie}.}
    \vspace{3pt}
    \label{tab:calvin}
    \small %
    \scalebox{0.9}{
    \begin{tabular}{l|c|c|ccccc|c}
    \toprule
    \multirow{2}{*}{Method} & \multirow{2}{*}{Type} & \multirow{2}{*}{\makecell[c]{Train\\episodes}}  & \multicolumn{5}{c|}{ Task completed in a row (\%) $\uparrow$} & \multirow{2}{*}{Avg. Len. $\uparrow$} \\
    &  &  & 1 & 2 & 3 & 4 & 5 & \\
    \midrule
    MCIL~\cite{lynch2020language} &\multirow{5}{*}{\shortstack{Language-conditioned\\Behaviour Cloning}} & All & 30.4 & 1.3 & 0.2 & 0.0 & 0.0 & 0.31\\
    HULC~\cite{mees2022hulc} & & All & 41.8 & 16.5 & 5.7 & 1.9 & 1.1 & 0.67 \\
    RT-1~\cite{brohan2022rt1} & & Lang & 53.3 & 22.2 & 9.4 & 3.8 & 1.3 & 0.90 \\
    RoboFlamingo~\cite{li2023roboflamingo} & & Lang & 82.4 & 61.9 & 46.6 & 33.1 & 23.5 & 2.48 \\
    GR-1~\cite{wu2023gr1} & & Lang & 85.4 & 71.2 & 59.6 & 49.7 & 40.1 & 3.06 \\
    \midrule
    3D Diffuser Actor~\cite{ke2024_3DDiffuserActor} & Diffusion Policy& Lang & 92.2 & 78.7 & 63.9 & 51.2 & 41.2 & 3.27 \\
    \midrule
    UniPi$^{*}$~\cite{du2024unipi} & \multirow{3}{*}{Planner + Executor} & All & 56.0	& 16.0	& 8.0	& 8.0	& 4.0	& 0.92 \\
    SuSIE~\cite{black202susie} & & All & 87.0 & 69.0 & 49.0 & 38.0 & 26.0 & 2.69 \\
    \baseline{\modelname (Ours)} & & \baseline{Lang} & \baseline{\textbf{96.0}} & \baseline{\textbf{83.5}} & \baseline{\textbf{70.8}} & \baseline{\textbf{57.5}} & \baseline{\textbf{45.4}} & \baseline{\textbf{3.53}} \\
    \bottomrule
    \end{tabular}
    }
    \vspace{-10pt}
\end{table}

\begin{table}[t]
    \centering
    \caption{\textbf{Performances with real-world robot tasks.} CLOVER achieves the best success rate and superior generalization capability across the board. }
    \label{tab:real_world}
    \vspace{3pt}
    \small %
    \scalebox{0.9}{
    \begin{tabular}{l | cccc | cc}
    \toprule
    \multirow{2}{*}{Method} & \multicolumn{4}{c|}{Long-horizon task} & \multicolumn{2}{c}{Single task}\\
    & Sub-task 1 & Sub-task 2 & Sub-task 3 & Avg. Len. $\uparrow$ & Pour shrimp & Stack bowls\\
    \midrule
    ACT~\cite{zhao2023learning} & 46.7 & 13.3 & 0.0 & 0.6  & 33.3 & 46.7\\
    R3M~\cite{nair2022r3m}  & 53.3 & 20.0 & 0.0 & 0.7 & 46.7 & 53.3\\
    RT-1~\cite{brohan2022rt1}  & 66.7 & 40.0 & 0.0 & 1.1  & \textbf{80.0} &66.7\\
    \midrule
    \baseline{\modelname (Ours)}  & \baseline{\textbf{93.3}} & \baseline{\textbf{86.7}} & \baseline{\textbf{26.7}} & \baseline{\textbf{2.1}} & \baseline{\textbf{80.0}} & \baseline{\textbf{86.7}}\\
    \bottomrule
    \end{tabular}
    }
    \vspace{-2pt}
\end{table}

\begin{figure}[t!]
    \centering
    \includegraphics[width=.9\linewidth]{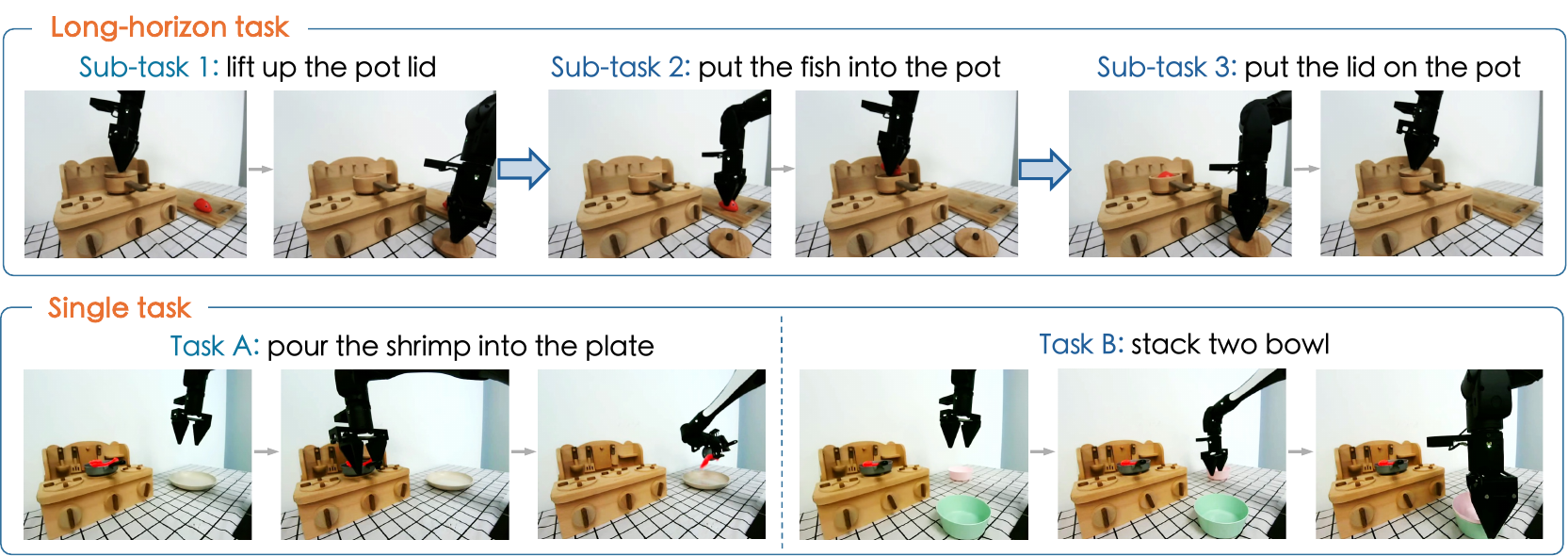}
    \vspace{-4pt}
    \caption{\textbf{Real-world robot setting.} We propose a long-horizon task encompassing three consecutive sub-tasks, where the failure of a prequel task will inevitably lead to failure of subsequent tasks. The additional single tasks are designed to validate the generalizability of CLOVER of all aspects.}
    \label{fig:real-world}
    \vspace{-7pt}
\end{figure}

\subsection{Experimental Setup} 
\label{sec:exp-setup}
\vspace{-5pt}

\noindent\textbf{Simulation tasks.} We conduct the majority of our experiments using CALVIN~\cite{mees2022calvin}, an evaluation benchmark designed for long-horizon, language-conditioned manipulation. CALVIN consists of four simulated environments (designated as A, B, C, and D), which differ in textures and object positions. Each environment comprises a Franka Emika Panda robot situated beside a desk equipped with various manipulable objects.
We train policy models on demonstrations collected from environments A, B, and C, and conduct zero-shot evaluations in environment D. The evaluation protocol involves assessing model performance on a comprehensive set of 1,000 unique instruction chains, each comprising five distinct tasks. The CALVIN benchmark provides an extensive dataset paired with natural language annotations, thereby facilitating the training of a generalized and reliable visual planner. 
Detailed implementation and training protocols are provided in \Cref{appendix:impl_details}.

\noindent\textbf{Real-world experiments.}
The real-robot experiments are conducted on the AIRBOT Play robotic arm. We propose a long-horizon task comprising three consecutive sub-tasks and two additional single tasks (``Pour shrimp into plate'' and ``Stack two bowls'', shown in \Cref{fig:real-world}). The fish and pot lid in sub-task 2 and sub-task 3, as well as the plate and bowl in two individual tasks, are randomly placed to reflect position generalizability. All metrics are evaluated with 15 independent runs.

\begin{table}[t]
    \centering
    \caption{\textbf{Generalization evaluation.} CLOVER excels under visual distractions and dynamic scenes, while the success rate of baselines dramatically drops. }
    \label{tab:real_world_generalization}
    \vspace{3pt}
    \small %
    \setlength{\tabcolsep}{3mm}{
    \scalebox{0.9}{
    \begin{tabular}{l|l | cccc }
    \toprule
    \multirow{2}{*}{Setting} & \multirow{2}{*}{Method} & \multicolumn{4}{c}{Long-horizon task} \\
    && Sub-task 1 & Sub-task 2 & Sub-task 3 & Avg. Len. $\uparrow$ \\
    \midrule
    \multirow{4}{*}{\makecell[l]{Visual\\Distraction}}&ACT~\cite{zhao2023learning} & 13.1 & 0 & 0 & 0.13\\
    &R3M~\cite{nair2022r3m}  & 20.0 & 0 & 0 & 0.20\\
    &RT-1~\cite{brohan2022rt1}  & 40.0 & 6.7 & 0 & 0.47\\
    &\baseline{\modelname (Ours)}  & \baseline{\textbf{73.3}} & \baseline{\textbf{66.7}} & \baseline{\textbf{6.7}} & \baseline{\textbf{1.47}}\\
    \midrule
    \multirow{2}{*}{\makecell[l]{Dynamic\\Scene}}&RT-1~\cite{brohan2022rt1}  & 33.0 & 0 & 0 & 0.33 \\
    &\baseline{\modelname (Ours)}  & \baseline{\textbf{80.0}} & \baseline{\textbf{53.3}} & \baseline{\textbf{20.0}} & \baseline{\textbf{1.53}}\\
    \bottomrule
    \end{tabular}
    }}
    \vspace{-6pt}
\end{table}

\subsection{Main Results}
\label{sec:exp-results}

\noindent\textbf{Visuomotor control on CALVIN.}
\Cref{tab:calvin} indicates that CLOVER achieves state-of-the-art performance on CALVIN, significantly outperforming previous methods with similar ``Planner + Executor'' pipelines. Without using gripper view images and proprio signals, our approach exceeds methods employing GPT-style transformers with pretraining, such as RoboFlamingo~\cite{li2023roboflamingo} and GR-1~\cite{wu2023gr1}. Note that all previous methods follow the CALVIN standard evaluation protocol~\cite{mees2022calvin}, where the simulator returns the signal that marks the completion of the current task. However, such task completion signals are not accessible in real-world environments. In fact, by leveraging the advantageous properties of our state embedding, we can determine the completion of a task autonomously without the signals. We leave this exploration to \Cref{appendix:illustration}.
\begin{wrapfigure}[19]{R}{0.35\textwidth}
    \centering
    \vspace{-8pt}
    \includegraphics[width=0.99\linewidth]{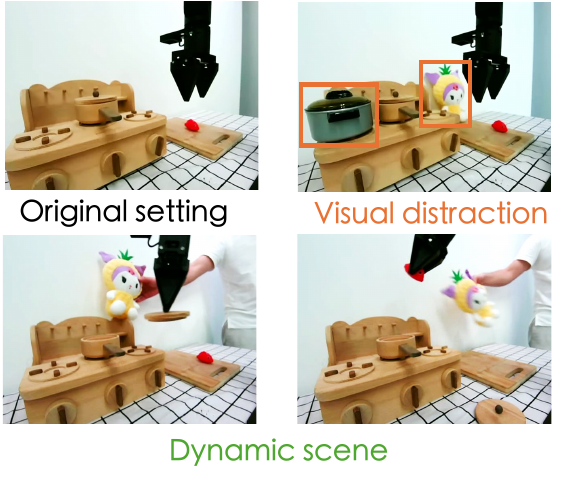}
    \vspace{-14pt}
    \caption{\small \textbf{Experiment setting of the generalization evaluation.} We place entirely new objects absent from training, alongside the interaction object to introduce \textcolor{orange}{visual distraction}. We test policies under \textcolor{myGreen}{dynamic conditions} by randomly placing and picking up a doll to create unpredictable visual changes.
    }
    \label{fig:generalization_exp_setting}
\end{wrapfigure}

\begin{figure}[t!]
    \centering
    \includegraphics[width=0.95\linewidth]{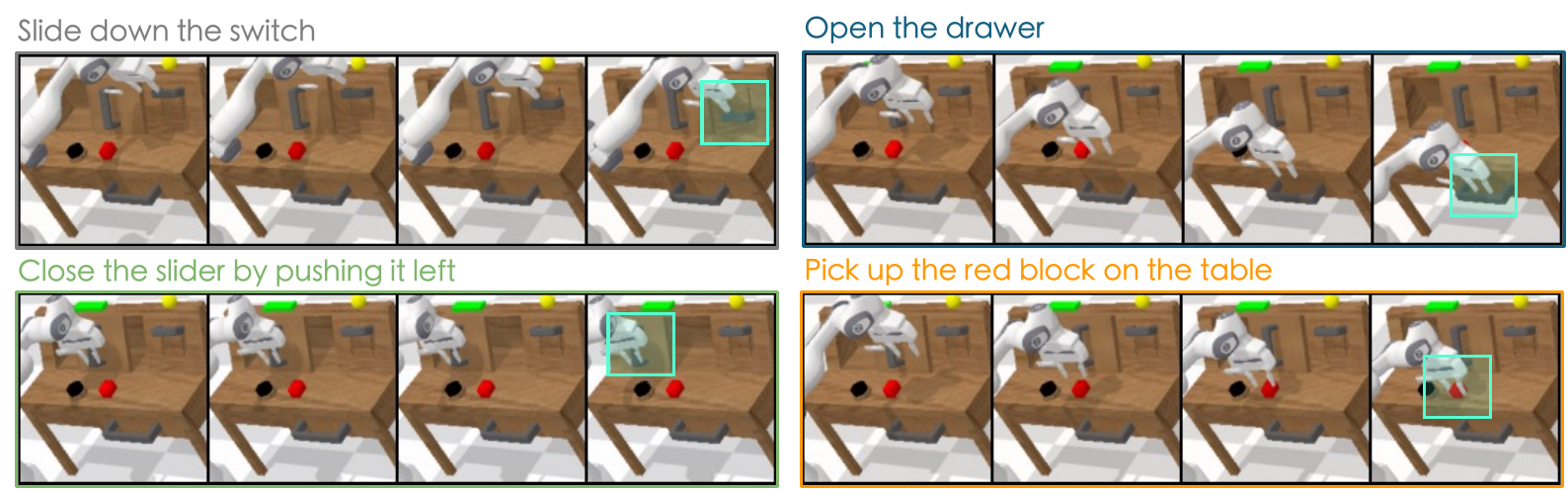}
    \vspace{-3pt}
    \caption{\textbf{Generated videos of diverse tasks conditioned on the same initial frame.} \modelname can generate precise visual plans corresponding to the tasks, facilitating low-level executor guidance. We downsample the video by 2 and exclude depth results in visualizations for simplicity.}
    \label{fig:vis_calvin_action_condition}
    \vspace{-5pt}
\end{figure}

\Cref{fig:vis_calvin_action_condition} shows our diffusion model's proficiency in instruction-following, with "Slide down the switch" as a representative task from the validation set and three others randomly proposed from the CALVIN's task pool. Our planner exhibits robust generalizability, producing reliable action trajectories for the subsequent policy. More visualizations are given in~\Cref{appendix:vis}.

\noindent\textbf{Manipulation with real-world robots.}
We present the evaluations of real-world robotic tasks in \Cref{tab:real_world}. CLOVER surpasses all baseline models by a considerable margin, especially on the long-horizon manipulation metric (+1.0 on Avg. Len.). Note that the lid knob is small and hard to grasp, which poses great challenges to policies' low-level precision. ACT~\cite{zhao2023learning} struggles to adjust the gripper to the right position before it should close
in the first task, while CLOVER doubles the success rate. Moreover, all three baselines we test fail on the last task, 
which requires the robot to re-cover the pot lid that was previously placed down in Task 1. 
In this scenario with high uncertainty, CLOVER shows a success rate of 26.7\%, indicating its stronger robustness and position generalization capability.

We further study the generalizability of CLOVER under visual distractions and dynamic environments, as inllustrated in~\Cref{fig:generalization_exp_setting}. Table~\ref{tab:real_world_generalization} lists the results of the experiment. CLVOER remains performant while manipulating under distractors, with the performance gap over baseline methods getting more pronounced. We provide qualitative analysis as shown in the Appendix~\Cref{fig:generalizability}. The visual planner effectively disentangles background distractions with foreground movements and generates appropriate plans. Additionally, the feedback-driven policy proves robust to dynamic scene variations, yielding better generalizability over our leading baseline method, RT-1.

\subsection{Discussion on Closed-loop v.s. Open-loop}
\label{sec:exp-close-vs-open}
\vspace{-5pt}

\noindent\textbf{Preliminaries.}
In this section, we conduct the ``open-loop'' experiments with the same diffusion and policy model, but do not incorporate the feedback mechanism in \Cref{alg:execution} to facilitate adaptive replan and sub-goal transitioning, which is a common practice in previous works~\cite{black202susie,ajay2024hip,du2024unipi}. 
\label{sec:open_vs_close}

\begin{figure}[t!]
    \centering
    \includegraphics[width=.98\linewidth]{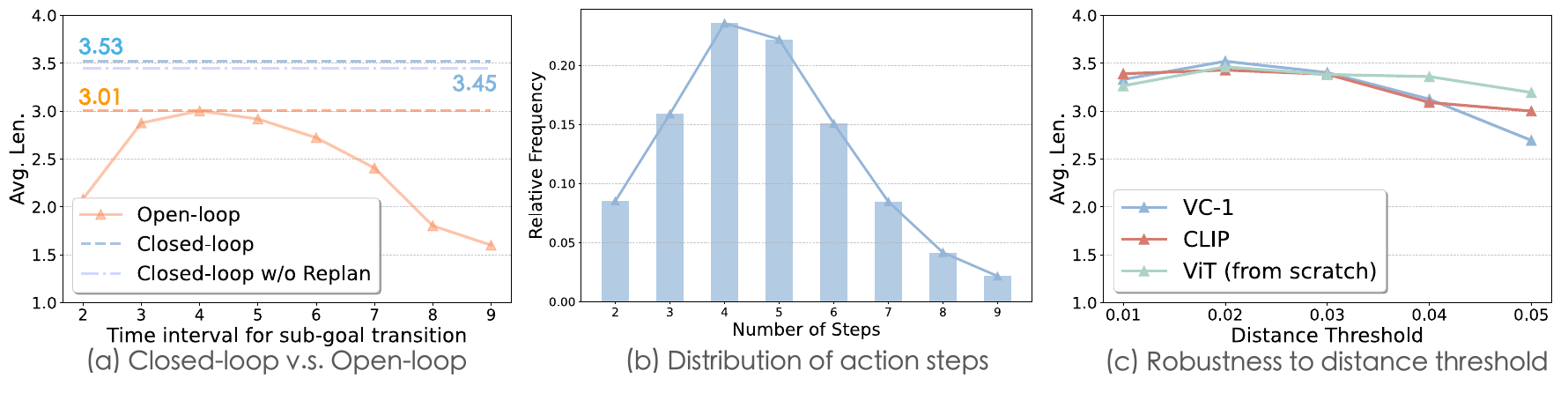}
    \vspace{-5pt}
    \caption{\textbf{Analysis and comparisons on closed-loop and open-loop roll-out on CALVIN.} (a) Comparative analysis of performance (Avg. Len.) through varying step lengths in open-loop control. Evaluations are conducted using identical models but employing different roll-out strategies. (b) The distribution of action steps taken in closed-loop roll-out to achieve each sub-goal. (c) Examination of the robustness of closed-loop control employing various visual encoders and distance thresholds. }
    \label{fig:closed_vs_open}
    \vspace{-10pt}
\end{figure}

\noindent\textbf{How does closed-loop work compared to open-loop?}
Figure \ref{fig:closed_vs_open}(a) compares the manipulation performance of open- and closed-loop execution. The proposed CLOVER, which adaptively selects sub-goals by assessing the distance between current observations and given sub-goals, demonstrates a significant performance improvement of +0.44 average completed task length on CALVIN. Incorporating adaptive replanning further boosts performance by detecting unreliable visual plans and preventing error propagation.
Notably, open-loop roll-out performance hinges on understanding the training specifics of both the planner and executor. Synchronizing the time interval for sub-goal transition with the frame intervals used during diffusion model training~($\Delta t = 5$ in our experiments) is needed for optimal performance. Figure \ref{fig:closed_vs_open}(a) illustrates the performance deterioration when these settings are not properly aligned. Its distribution pattern aligns with the closed-loop roll-out step distribution in Figure \ref{fig:closed_vs_open}(b), supporting the necessity of adaptive steps for optimal performance.


\begin{wraptable}[10]{R}{0.5\textwidth}
    \centering
    \small
    \vspace{-12pt}
    \caption{\textbf{Error measurments with different representations.} 
    Our method shows exceptional cross-tasks robustness on CALVIN benchmark.
    }
    \scalebox{0.87}{
    \begin{tabular}{l|ccccc|c}
    \toprule
    \multirow{2}{*}{Method} &\multicolumn{5}{c|}{ Task completed in a row (\%) $\uparrow$} & \multirow{2}{*}{\makecell[c]{Avg.\\Len.} $\uparrow$} \\
     & 1 & 2 & 3 & 4 & 5 &  \\
    \midrule
    CLIP~\cite{radford2021clip}& 72.4	&46.8	&25.0	&13.7	&5.1	&1.63\\
    LIV~\cite{ma2023liv} & 70.8	&48.2	&29.2	&18.2	&10.2	&1.77\\
    \midrule
    \baseline{\modelname} & \baseline{\textbf{96.0}} & \baseline{\textbf{83.5}} & \baseline{\textbf{70.8}} & \baseline{\textbf{57.5}} & \baseline{\textbf{45.4}} & \baseline{\textbf{3.53}}\\
    \bottomrule
    \end{tabular}
    }
    \label{tab:measurement}
\end{wraptable}
\noindent\textbf{How generalizable is our feedback mechanism?} We investigate the effect of different visual encoders across varying distance thresholds ($D_{S}$ in \Cref{alg:execution}) used in policy models, as depicted in \Cref{fig:closed_vs_open}(c). A larger $D_{S}$ indicates a higher error tolerance.
It can be seen that there is a consistent pattern across different distance thresholds for all encoders, with peak performance observed at $D_{S}=0.02$. Adopting VC-1~\cite{vc2023} demonstrates the highest performance, while training a ViT-Base~\cite{dosovitskiy2020image} encoder from scratch yields exceptional stability, with the lowest result being 3.19. The results manifest the robustness of our feedback mechanism which is independent of specific encoders and does not require customized hyperparameters for different policy models.
We compare the robustness of our error measurement scheme against other representations, specifically the dense reward learning framework LIV~\cite{ma2023liv} and CLIP features~\cite{radford2021clip}. We maintain the same model for both the visual planner and the low-level policy, with the exception of the measurements used to determine sub-goal transitions and replanning.
Our findings (as in~\Cref{fig:explain_closed_loop}) indicate that CLIP features and LIV exhibit a narrower range of values, prompting us to set $D_{S}$ to $2e-3$. The results presented in \Cref{tab:measurement} show that unreliable measurements can lead to performance that is even worse than in the open-loop setting. Moreover, we do not incorporate additional contrastive objectives as in LIV, but investigate the inherent properties of the inverse dynamics-based policy.

\subsection{Ablation Studies}
\label{sec:exp-ablation}
\vspace{-5pt}

\noindent\textbf{Ablations on the diffusion model~(visual planner).} 
We conduct a comparative analysis of the high-level plan (video generation) quality. Results presented in \Cref{tab:abl_diffusion} reveal that our method outperforms AVDC~\cite{Ko2023Learning} across all video generation metrics~\cite{fan2019metrics}, both of which are built upon the Imagen framework~\cite{saharia2022imagen}, alleviating the learning and generalization burden on the policy model. Notably, AVDC struggles to generate visual plans consistent with the task description, resulting in significant performance degradation on the CALVIN benchmark. Additionally, the optical flow-based regularization not only brings a generation quality improvement across all aspects, but also significantly accelerates the training convergence. Please refer to~\Cref{supp:abl} for further analysis. 

\begin{table}[t]
    \centering
    \caption{\textbf{Comparison on video prediction.} With optical flow-based regularization and architectural modifications, CLOVER generates videos of higher quality and more accurately.
    $^\dag$: Reproduced.}
    \vspace{3pt}
    \label{tab:abl_diffusion}
    \small %
    \scalebox{0.85}{
    \begin{tabular}{l|ccccc|c}
    \toprule
    Method & SSIM $\uparrow$	& PSNR $\uparrow$	& LPIPS $\downarrow$ & FID $\downarrow$ & RMSE~(Depth) $\downarrow$ & Avg. Len. (CALVIN) $\uparrow$ \\
    \midrule
    AVDC$^\dag$~\cite{Ko2023Learning} &0.837 & 20.76 & 0.086 & 12.74 & - & 1.42\\
    \modelname (w/o Flow Reg.) & 0.848 & 21.42 & 0.076 & 12.38 & 0.084 & 3.26 \\
    \baseline{\modelname (Ours)} & \baseline{\textbf{0.858}} & \baseline{\textbf{22.19}} & \baseline{\textbf{0.062}} & \baseline{\textbf{12.00}} & \baseline{\textbf{0.063}} & \baseline{\textbf{3.53}} \\
  
    \bottomrule
    \end{tabular}
    }
    \vspace{-2pt}
\end{table}

\begin{figure}[t!]
    \centering
        \includegraphics[width=\linewidth]{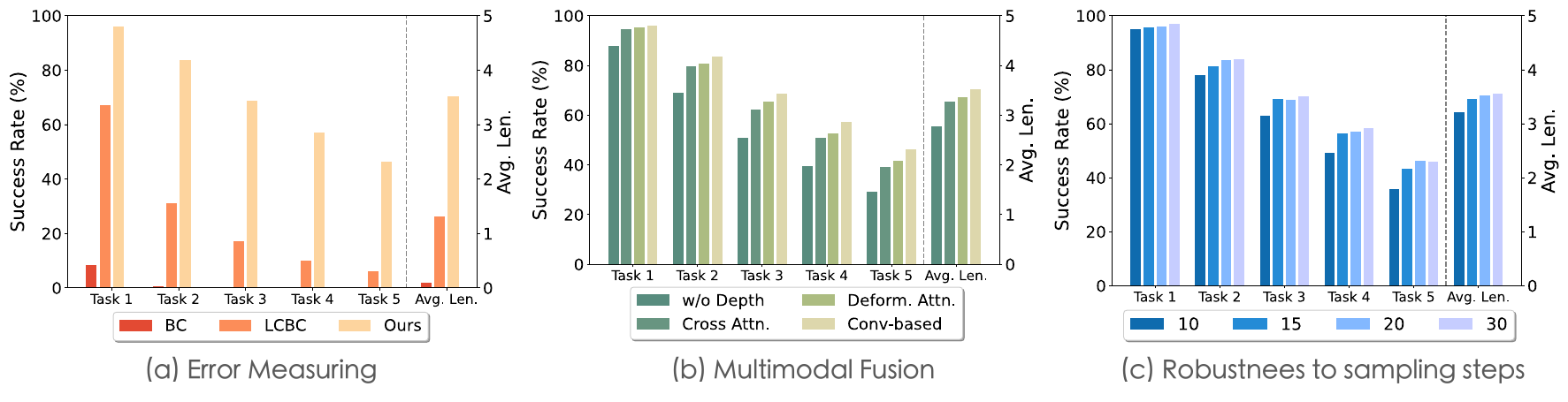}
    \vspace{-15pt}
    \caption{\textbf{Ablations on the policy model:} (a) error measuring mechanism and (b) multimodal fusion module, with discussions on (c) robustness to different sampling steps when generating sub-goals. We report success rates and
    the average length of completed tasks (divided by dash lines in each plot).}
    \label{fig:abl_policy}
    \vspace{-10pt}
\end{figure}

\noindent\textbf{Ablations on the policy model.} Besides the ablation of feedback mechanisms in \Cref{sec:exp-close-vs-open}, we additionally evaluate the performance of CLOVER using various error measurement approaches, illustrated in Figure \ref{fig:abl_policy}(a). Policies learned with behavior cloning (BC, \Cref{fig:explain_closed_loop}(b)) and LCBC~\cite{walke2023bridgedata} serve as baselines that do not employ error measurement, while CLOVER exceeds them by a notable margin. \Cref{fig:abl_policy}(b) demonstrates that incorporating geometry information from depth data leads to a not trivial improvement on CALVIN, with convolution-based multimodal fusion modules achieving the best performance. We also examine the robustness of our policy to generation quality by varying the sampling steps of the diffusion model. As shown in \Cref{fig:abl_policy}(c), increasing the sampling steps generally provides generated videos with more details but shows diminishing returns. We set the sampling step to 20 to strike a balance between performance and efficiency.

\section{Conclusion}
\label{sec:conclusion}
\vspace{-5pt}
In this paper, we present a generalized closed-loop visuomotor control framework, termed CLOVER. It comprises a visual planner that specifies desired sub-goals, a policy that executes actions as planned, and a feedback-driven control strategy to realize long-horizon robotic tasks. CLOVER excels in both simulation and real-world applications, showcasing the virtue of our feedback mechanism.

\noindent\textbf{Limitation and future works.} 
We validate \modelname for simulation and real-world experiments by training the models heavily on the corresponding data. However, emerging evidence suggests both the diffusion models and IDM-based policies exhibit out-of-distribution generalizability~\cite{blattmann2023svd,brandfonbrener2024inverse,chi2023diffusion}. Visual planner can be trained with actionless videos, and IDM can be learned data-efficiently with random actions with corresponding observations. This points to the potential of our framework for performing few-shot and long-horizon manipulations by pre-training on web-scale datasets. 

\section*{Acknowledgments}
This work was supported by National Key R\&D Program of China (2022ZD0160104), NSFC (62206172), Shanghai Committee of Science and Technology (23YF1462000), and China Postdoctoral Science Foundation (2023M741848). We  thank DISCOVER Robotics for providing the hardware consulting used in this research as well.

{
\small
\bibliographystyle{unsrt}
\bibliography{bibliography_short, bibliography_custom}
}

\clearpage
\appendix

\section*{
\Large{\textit{Supplementary}}
}
\section{Examples of Test-time Execution with Error Measurement}
\label{appendix:illustration}
\vspace{-5pt}

In this supplemental section, we provide examples of how our proposed deviation quantification helps the autonomous test-time execution. In particular, as mentioned in \Cref{sec:feedback}, it can be adopted for replanning when sub-goals are infeasible or unrealistic, and transitioning to the next desired state when the robot reaches each sub-goal. In the meantime, we introduce the application of task completion assessment, though not utilized directly for performance comparisons in CALVIN. 

\begin{figure}[h]
    \centering
    \vspace{-2pt}
    \includegraphics[width=0.92\linewidth]{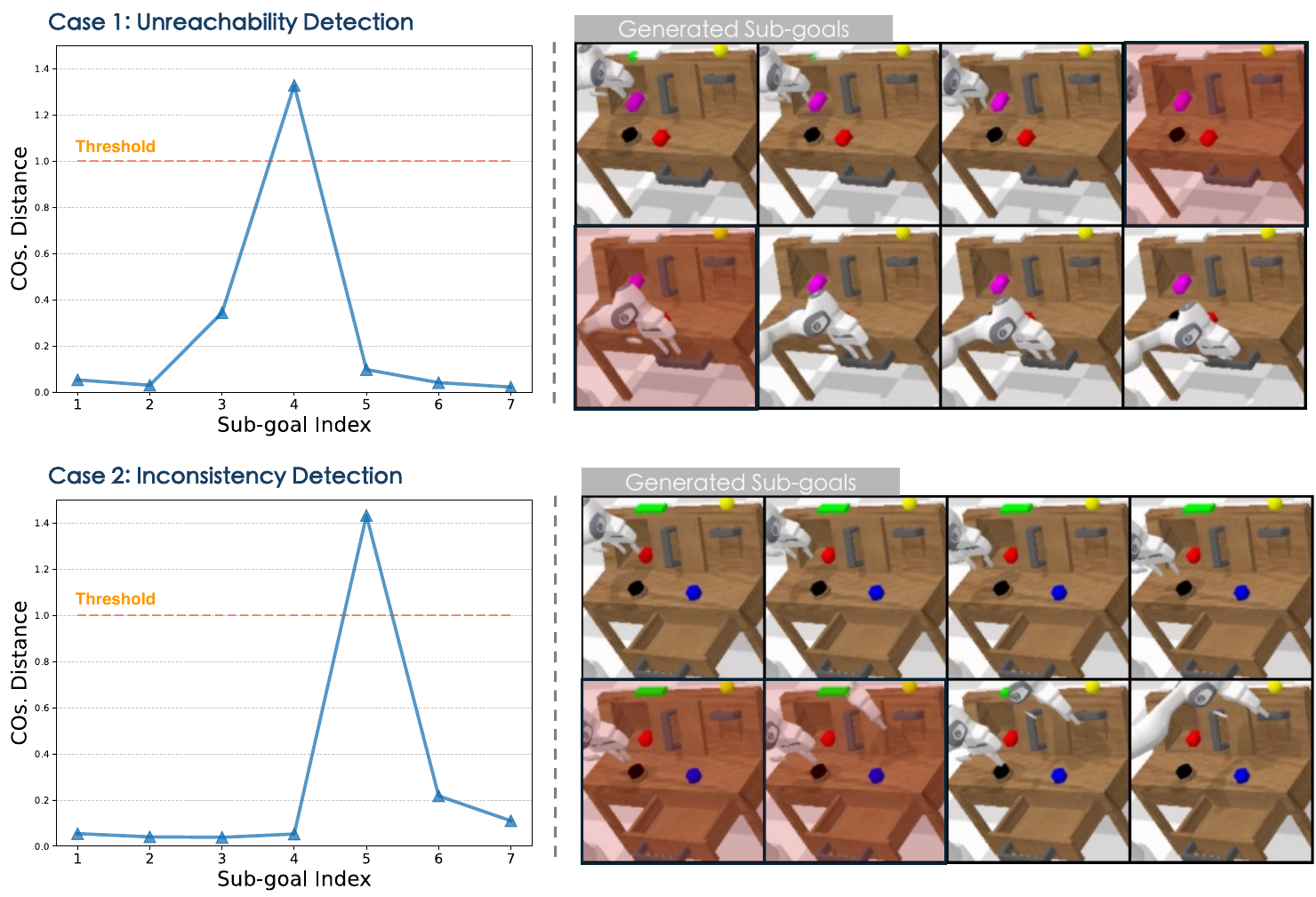}
    \vspace{-5pt}
    \caption{\textbf{Automatic identification of unreliable sub-goals generated.} We set the distance threshold for replanning as $D_{R}=1.0$ under all circumstances. Ideally, the distances between generated frames remain consistent and relatively small, whereas significant variations occur during unstable generation. Examining the distance within two adjacent frames, we can detect erroneous plans generated by the diffusion model before passing them to the subsequent policy model for execution.}
    \label{supp_fig:vis_inconsistency}
\end{figure}

\noindent\textbf{Sub-goal replan.} We first show how to identify unreachable sub-goals to mitigate the uncertainty of diffusion-based visual planners. \Cref{supp_fig:vis_inconsistency} gives examples of common cases showing the instability of the diffusion model. We measure the cosine distance between the state embeddings of consecutive frames generated by the model. Specifically, consistent and reasonable plans exhibit a stable intra-frame distance, typically below 0.2, whereas inconsistencies are characterized by a significant increase, often exceeding 1. Therefore, we set the distance threshold for replanning as $D_{R}=1.0$ under all circumstances, which proves to work effectively in practice. The distinctive variation verifies the sensitivity and virtue of the measuring ability of state embeddings. By adaptively detecting erroneous plans and reinitializing generation, we prevent error propagation to the subsequent policy.

\begin{figure}[t]
    \centering
    \includegraphics[width=0.95\linewidth]{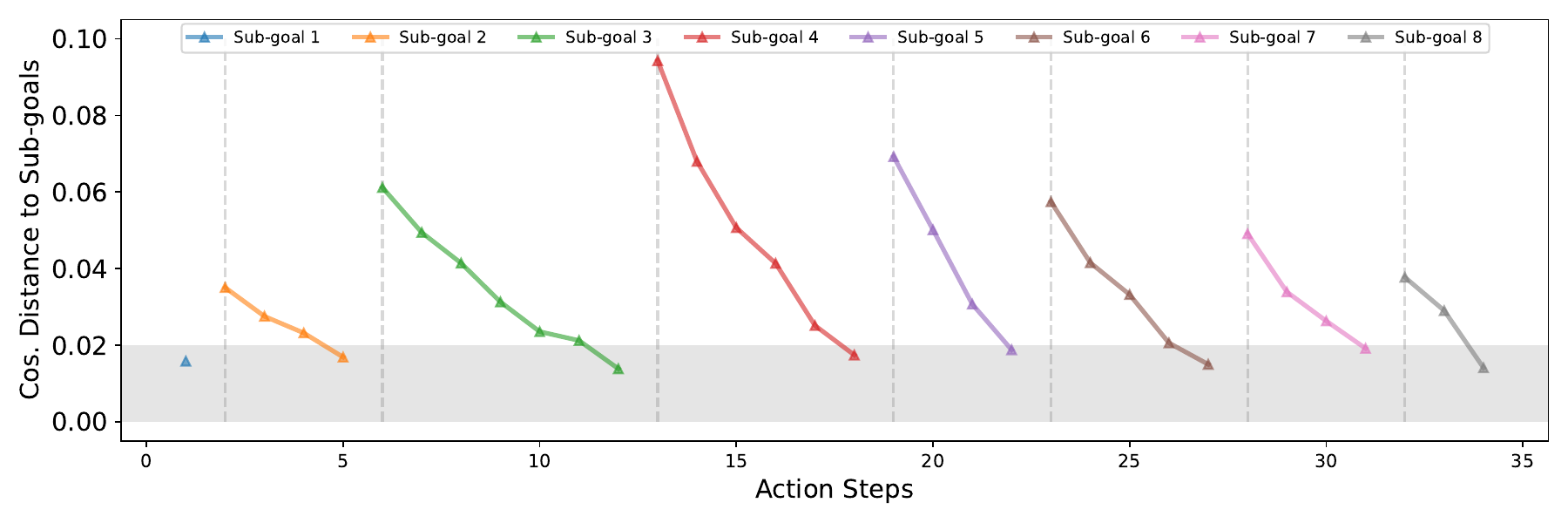}
    \vspace{-8pt}
    \caption{\textbf{Adaptive sub-goal transitions.} The cosine distance between the current observation and the selected sub-goals is plotted, with dashed gray lines indicating the transitions between sub-goals. The distance threshold $D_{S}$ for sub-goal transitioning is set to 0.02. Our policy effectively reaches each assigned sub-goal and minimizes errors through an adaptive number of action steps.}
    \label{fig:adaptive_sub_goal_transition}
    \vspace{-10pt}
\end{figure}

\noindent\textbf{Sub-goal transition.} Distinguished from previous works that adopt fixed action steps mentioned in \Cref{sec:exp-close-vs-open}, our framework incorporates a predefined distance threshold $D_{S}$ (refer to~\Cref{alg:execution}) to facilitate adaptive transitioning between sub-goals upon achievement. In~\Cref{fig:adaptive_sub_goal_transition}, we illustrate the distance variation between the current observation and selected goals during the roll-out process. In our experimental setup, \modelname dynamically adjusts the number of steps for approaching each sub-goal, ranging from 1 to 7, and transitions to the next sub-goal once the distance falls below the threshold $D_{S}$. Notably, the distribution of generated goals in the latent space is uneven, and the policy does not adhere to a fixed speed in reaching targets. This is evident from the varying starting distances and changing slopes of the approach process for each sub-goal in \Cref{fig:adaptive_sub_goal_transition}. These observations further underscore the necessity of a feedback mechanism to enable the adaptive sub-goal transition to mitigate error accumulation.

\begin{figure}[t]
    \centering
    \includegraphics[width=0.95\linewidth]{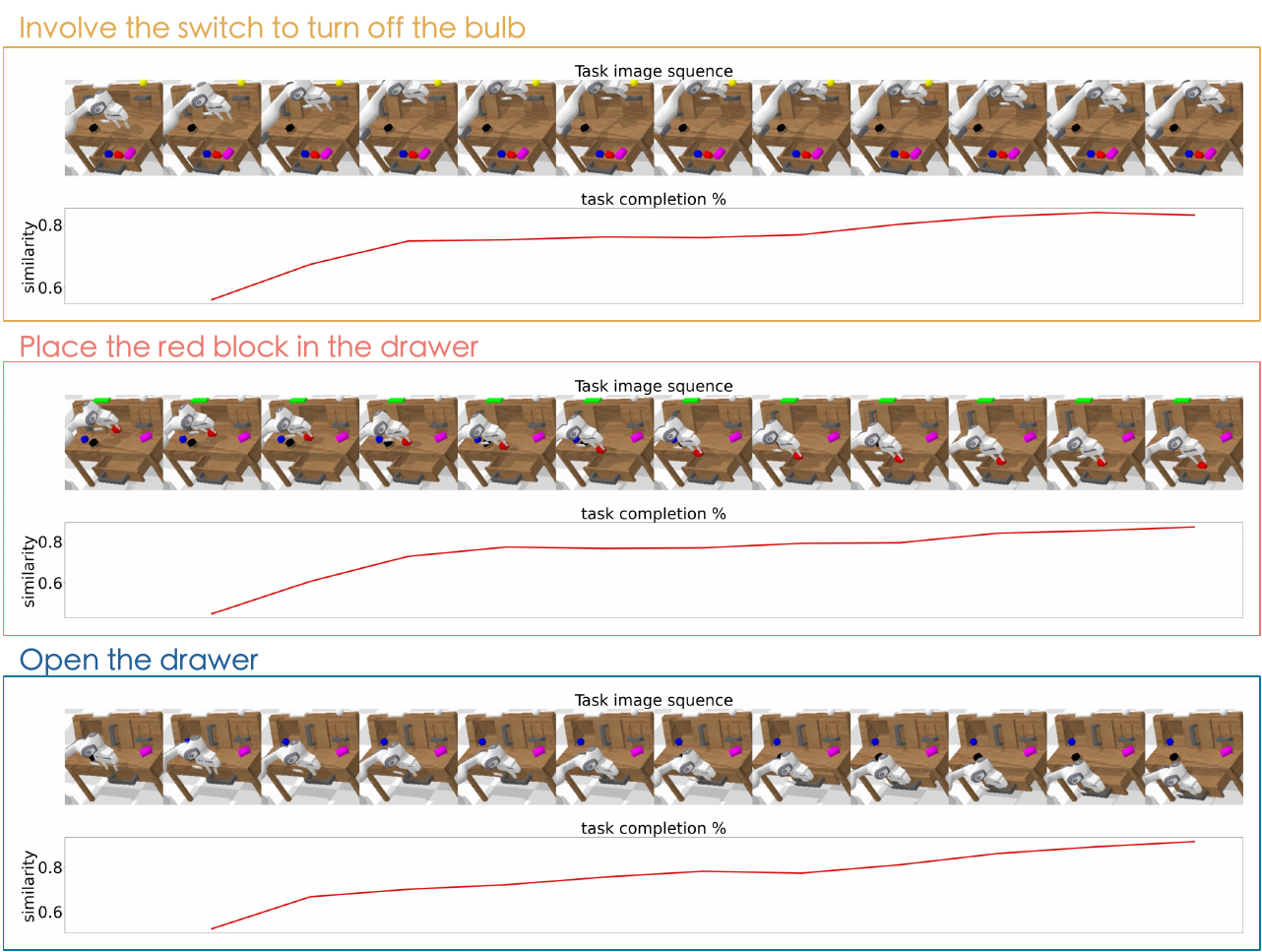}
    \vspace{-2pt}
    \caption{\textbf{Task completion score given by our value function.} As the robot approaches each subgoal and eventually completes the task described in the text, the learned value function monotonically increments, which could be helpful for task completion assessment.}
    \label{fig:task_reward}
\end{figure}

\noindent\textbf{Task completion assessment.} All previous methods follow the standard evaluation protocol of CALVIN~\cite{mees2022calvin}, where the simulator updates the observation after each action roll-out and returns a signal marking the completion status of the current task. The long-horizon manipulation tasks in CALVIN consist of a sequence of five consecutive sub-tasks. Consequently, the completion status of the current sub-task serves as a valuable signal for policies to progress to the subsequent one. However, such signals are not accessible in real-world environments. 

To fully exploit the advantageous properties of our state embedding as introduced in~\Cref{sec:executor}, we also explore using them to facilitate end-of-task assessment. Since we model latent actions for each step with the deviation of current and sub-goal embeddings, 
it is natural to exploit the deviation between the embeddings of the task's initial and final states to represent the latent action for the entire task. Inspired by LIV~\cite{ma2023liv}, we introduce a contrastive objective for language and state embeddings to establish a dense (value) reward, where a high reward for the current state represents an approaching status of the entire task. Specifically, we use the subtraction of the randomly sampled intermediate state and initial state embeddings as the negative example, while the subtraction of final and initial embeddings as the positive example. We then compute the InfoNCE loss~\cite{oord2018representation} with the encoded text goal to align image and language goals. Intuitively, a current state closing to the ground-truth final state resembles the overall text descriptions. Notably, the training process is built upon the pretrained policy and text encoders, with only a lightweight projector trained to align modalities.
Figure~\ref{fig:task_reward} presents examples of using this reward design to assess a task's completion status. Such characteristic demonstrates its potential as a task completion judge.

\begin{figure}[h]
    \centering
    \includegraphics[width=0.99\linewidth]{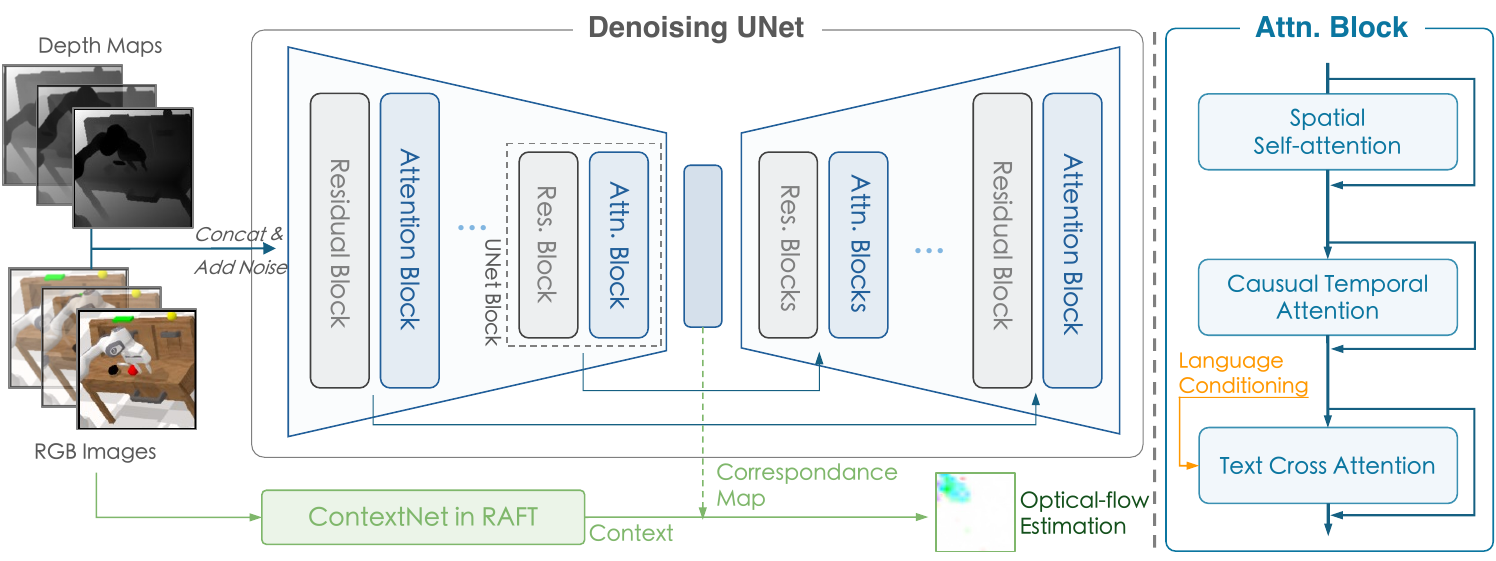}
    \vspace{-3pt}
    \caption{\textbf{The architecture of our visual planner.} We augment the original UNet proposed in Imagen~\cite{saharia2022imagen} with casual temporal attention to improve intra-frame consistency and additional cross-attention-based language conditioning. Combining a lightweight ContextNet introduced in RAFT~\cite{teed2020raft}, we estimate optical flows with a correspondence map of diffusion latent embeddings. }
    \label{fig:planner_arch}
    \vspace{-6pt}
\end{figure}

\section{Implementation Details}
\label{appendix:impl_details}
\vspace{-5pt}

\subsection{Model Architecture}
\label{appendix:arch_details}
\vspace{-5pt}

\noindent\textbf{Visual planner.}
\Cref{fig:planner_arch} depicts the detailed architecture of our visual planner. Following~\cite{du2024unipi,Ko2023Learning}, we adopt a video diffusion model to generate synthesized sub-goals. Considering diffusion models are generally computationally expensive, we down-scale the model size to achieve a balance between the efficiency for robot manipulation tasks and high fidelity for video generation.

The visual planner needs to generate temporally coherent and consistent videos (sub-goals) to guide the subsequent executor. In line with previous studies on video generation~\cite{blattmann2023svd, Ko2023Learning, yang2024generalized}, we concatenate the input condition frame $O_{0}$ to all the future frames $O_{1:K}$ to ensure coherency. Furthermore, to foster better temporal consistency and causal reasoning ability, causal temporal attention is adopted to encourage the full exploitation of historical information for predicting future interactions. Combining this with the factorized spatial-temporal convolution~\cite{sun2015human} in AVDC~\cite{Ko2023Learning}, our visual planner can faithfully reason about temporal causalities with improved training and inference efficiency.

To perform optical flow estimation $\hat{F}_{k \rightarrow k+1}$ with diffusion latents we employ an additional context network to encode context information from the $k$-th frame following RAFT~\cite{teed2020raft}. In contrast to utilizing an external flow estimation model directly~\cite{Ko2023Learning,lai2018learning}, our approach incurs a minimal parameter overhead of 0.7M during training, with the optical flow estimation component being discarded during test-time control. 

\noindent\textbf{Feedback-driven policy.}
We leverage the off-the-shelf pre-trained visual encoders (\textit{i.e.}, VC-1~\cite{vc2023}, CLIP~\cite{radford2021clip}, and DINO~\cite{oquab2023dinov2}) to imbue the training process with visual priors. As for depth, we intuitively adopt a transformer-based encoder (ViT-Small~\cite{dosovitskiy2020image}) to align multi-modal information in the latent space. To ensure compatibility with various pre-trained visual encoders, the input and patch size of the depth encoder are adapted to be consistent with the visual token length, thereby facilitating the widespread adoption of our policy. For fusing RGB ($\mathbf{F}_{\text{RGB}} \in \mathbb{R}^{l \times d_{\text{RGB}}}$) and depth features ($\mathbf{F}_{\text{Depth}} \in \mathbb{R}^{l \times d_{\text{Depth}}}$), we simply concatenate them on channels followed by a 3$\times$3 convolutional layer to integrate spatial local information and reduce the channel dimension to $d_{\text{RGB}}$. It is then followed by a Squeeze-and-Excitation module~\cite{hu2018squeeze} as channel attention to select important fused features. 

For the subsequent token aggregator, we set the number of attention heads~\cite{vaswani2017attention} to 12 in the multi-head attention module, and multiply the channel width by 4 in the MLP to enhance the representation capacity. The action decoding head is a three-layer MLP with ReLU as the activation function.

\subsection{Training Protocol}
\label{appendix:train_details}
\vspace{-5pt}

\noindent\textbf{Visual planner.} Our diffusion model-based planner can be factorized as $p_{\phi}(O_{1:K}\ | \  O_{0}, c_{l})$, with $c_{l}$ presenting the condition given by the language. During the training process, it acts as a denoising function $\epsilon_{\phi}$ predicting noises applied on future video frames $O_{1:K}$~\cite{ho2020denoising}. Given the noise scheduling $\beta_{t}$, the training objective of the diffusion model is:
\begin{equation}
    L_{\text{diff}} = \frac{1}{K} \sum_{k=1}^{K} \Vert \epsilon - \epsilon_{\phi} ( \sqrt{1 - \beta_{t}} \cdot O_{k} + \sqrt{\beta_{t}} \cdot \epsilon\  | \  t , c_{l})\Vert^{2},
\end{equation}
where the noise $\epsilon \in \{\epsilon_{\text{RGB}}, \epsilon_{\text{Depth}} \}$ is drawn from a multivariate standard Gaussian distribution, and $t$ represents a randomly selected diffusion step. Specifically, we sample noises separately for RGB and depth from two independent distributions. We further adopt the min-SNR weighting strategy~\cite{hang2023efficient} to speed up convergence. Combining the flow-based regularization term in \Cref{sec:planner}, the final optimization objective of the visual planner can be formulated as:
\begin{equation}
    L_{\text{planner}} = L_{\text{diff}} + \lambda L_{\text{reg}},
\end{equation}
where $\lambda$ is a balancing factor and is set to 0.1 by default. In our experiments on the CALVIN~\cite{mees2022calvin} benchmark, we train the diffusion model for 300k iterations with a learning rate of 1e-4. Models are trained on a system equipped with 8 A100 GPUs with the batch size set as 32. We adopt the AdamW optimizer without weight decay. Besides, we track an exponential moving average (EMA) of the model parameters with a decay rate of 0.999 and use the EMA parameters at test time. For real-world experiments, we tune the diffusion model for 50,000 iterations on 50 collected demonstrations. Due to hardware limitations, we are not able to collect depth data in a real environment, so the model generates RGB images only.

For test-time execution, the DDIM sampler~\cite{song2020denoising} is employed, with 20 sampling steps to strike a balance between efficiency and quality. The text guidance weight is set to 4 for generating visual plans that align with linguistic descriptions.

\noindent\textbf{Feedback-driven policy.} To optimize the policy model, we leverage mean squared error and binary cross-entropy loss to supervise the end-effector's position $a_{\text{EE}} \in \mathbb{R}^{6}$ and gripper state $a_{\text{gripper}} \in \mathbb{R}^{1}$, respectively. In each training episode, two frames with an interval ranging from 1 to $k_{\text{max}} = 5$ are sampled as inputs to enhance the model's robustness. We train the policy on ABC training split of CALVIN for 10 epochs with a batch size of 128. Only the relative cartesian action of a single timestamp is used for training. The training process takes around 10 hours on 8 A100 GPUs.

\section{Extended Ablation Studies}
\label{supp:abl}
\vspace{-5pt}

\begin{figure}[t]
    \centering
        \centering
        \includegraphics[width=0.95\linewidth]{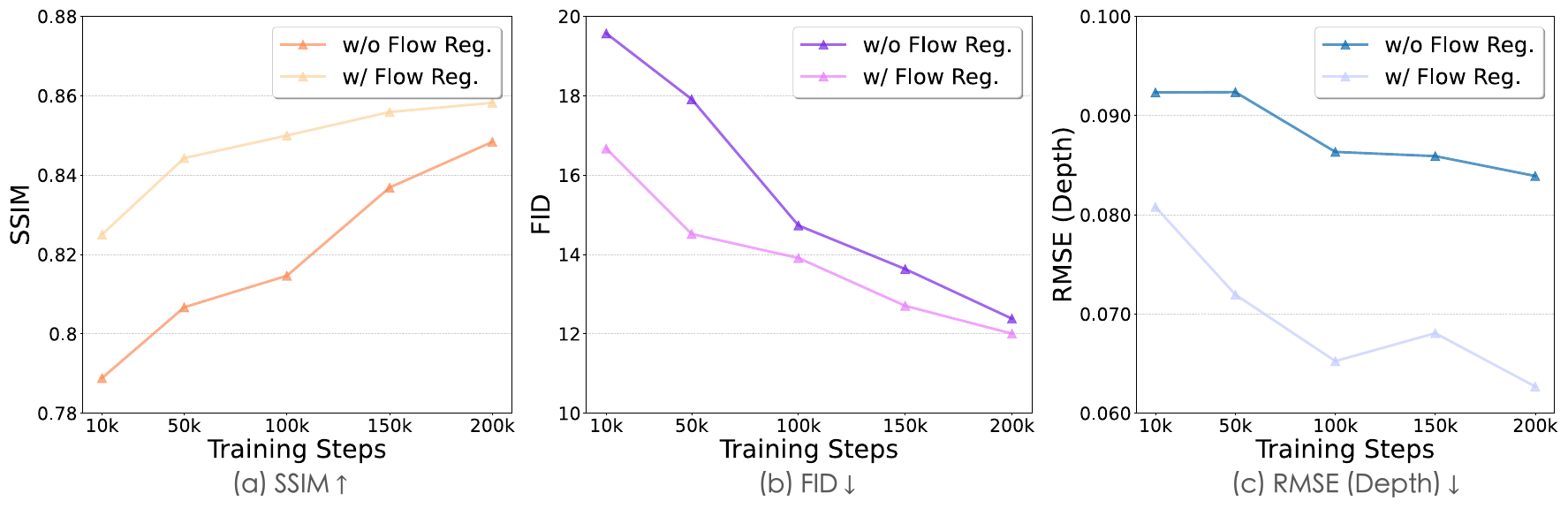}
    \vspace{-5pt}
    \caption{\textbf{Quantitative performance of video diffusion model with and without the flow-based regularization term.} The optical flow-based regularization endows the video diffusion model with more efficient training convergence and notable performance improvement on CALVIN.}
    \label{fig:abl_diffusion}
\end{figure}

\begin{table}[t]
\small
    \centering
    \caption{\textbf{Modeling Inverse Dynamics with different visual encoders.} CLVOER achieves promising results on CALVIN with different visual encoders of varying sizes. $^{*}$: Trained from scratch.}
    \vspace{5pt}
    \label{tab:abl_encoder}
    \small %
    \scalebox{0.9}{
    \begin{tabular}{l|c|ccccc|c}
    \toprule
    \multirow{2}{*}{Encoders}& \multirow{2}{*}{Params.}& \multicolumn{5}{c|}{ Task completed in a row (\%) $\uparrow$} & \multirow{2}{*}{Avg. Len. $\uparrow$}  \\
    & & 1 & 2 & 3 & 4 & 5 & \\
    \midrule
    ViT-S$^{*}$~\cite{dosovitskiy2020image} &22M&96.0	&80.2	&64.4	&52.6	&41.4	&3.35\\
    ViT-B$^{*}$~\cite{dosovitskiy2020image} &86M&\textbf{96.6}	&\textbf{83.8}	&67.6	&55.0	&43.2	&3.46\\
    CLIP~\cite{radford2021clip}        &86M&94.0	&81.2	&68.0	&56.2	&43.4	&3.43\\
    DINOv2~\cite{oquab2023dinov2}      &307M&94.8	&80.4	&67.0	&55.4	&43.8	&3.41\\
    \midrule
    \baseline{VC-1~\cite{vc2023}} & \baseline{86M} & \baseline{96.0} & \baseline{83.5} & \baseline{\textbf{70.8}} & \baseline{\textbf{57.5}} & \baseline{\textbf{45.4}} & \baseline{\textbf{3.53}}  \\
    \bottomrule
    \end{tabular}
    }
\end{table}
\begin{figure}[t!]
    \centering
    \includegraphics[width=0.95\linewidth]{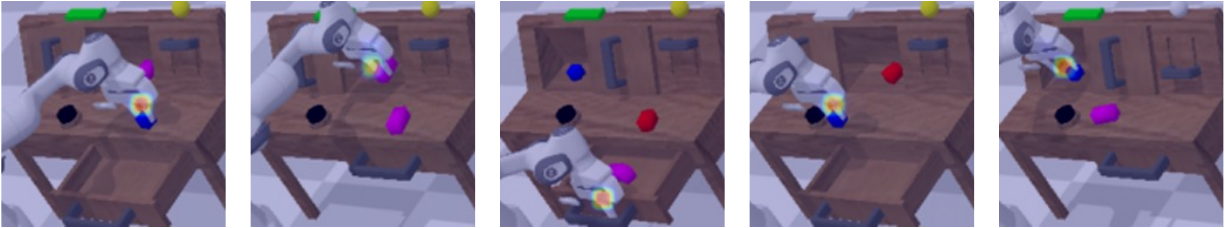}
    \caption{\textbf{Visualization on attention maps of the policy model.} The policy model demonstrates the ability to direct attention towards the end-effector and the object it interacts with, without the need for explicit supervision during the learning process.}
    \label{supp_fig:attn_map}
    \vspace{-4pt}
\end{figure}

\noindent\textbf{Optical flow based regularization.} We investigate the effect of optical flow-based regularization in model convergence, besides the results reported in \Cref{tab:abl_diffusion}. 
As depicted in \Cref{fig:abl_diffusion}, our analysis illustrates the impact on model performance across different training phases. The regularization significantly boosts the model's convergence, evidenced by consistently improved metrics. Notably, the visual planner trained with 10k iterations outperforms its 100k iterations counterpart in terms of SSIM and RMSE. Furthermore, the final model, trained over 200k iterations, demonstrates superior video quality with a 0.38 decrease in FID and 0.02 in RMSE. By leveraging the pixel-space correlation to regularize the diffusion model's latent space representations, the regularization term encourages the model to focus on movement and interaction dynamics, leading to improved convergence and performance on manipulation videos.

\noindent\textbf{Generalization to different visual encoders.} In \Cref{tab:abl_encoder}, we report detailed statistics regarding experiments of different visual encoders for our feedback mechanism in \Cref{fig:closed_vs_open}(c), under the default distance threshold setting. 
Notably, training a 22M parameters ViT-Small model~\cite{dosovitskiy2020image} from scratch already yields satisfactory results. VC-1~\cite{vc2023}, which is trained on tons of robotics data, provides the policy with a better visual prior and brings the best performance. In general, we hypothesize that the performance bottleneck of the framework still lies in the visual planner, while IDM-based policies are easier to learn and generalize.

\section{Extended Visualizations}
\label{appendix:vis}
\vspace{-5pt}

\noindent\textbf{Attention maps of our policy.} As discussed in \Cref{sec:feedback}, the state embeddings extracted by the token aggregator distribute reasonably in the latent space and exhibit decent measurement capabilities. We plot attention maps that show the tokens chosen by the token aggregator of the policy model in \Cref{supp_fig:attn_map} to provide additional explanation for the observation. As visualized, our model focuses on information essential for action execution, specifically the end-effector and the object it interacts with. One thing worth noting is that it is not explicitly supervised with any form of affordance.

\begin{figure}[H]
    \centering
    \includegraphics[width=0.9\linewidth]{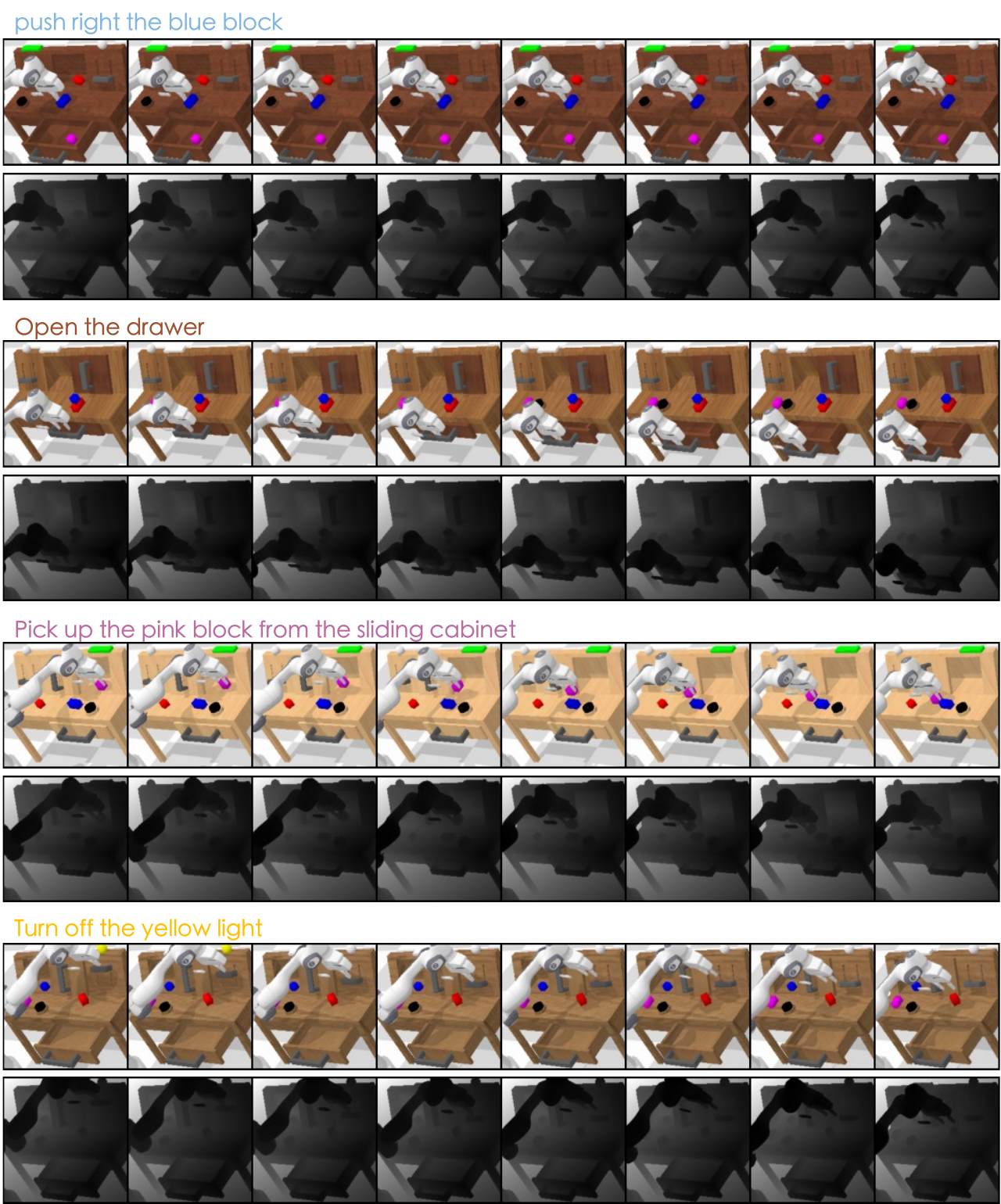}
    \caption{\textbf{Visualization on generated RGB-D visual plans.} Our model can generate reliable sub-goals with high consistency between RGB and depth.}
    \label{supp_fig:visualization_full}
\end{figure}

\begin{figure}[H]
    \centering
    \includegraphics[width=0.9\linewidth]{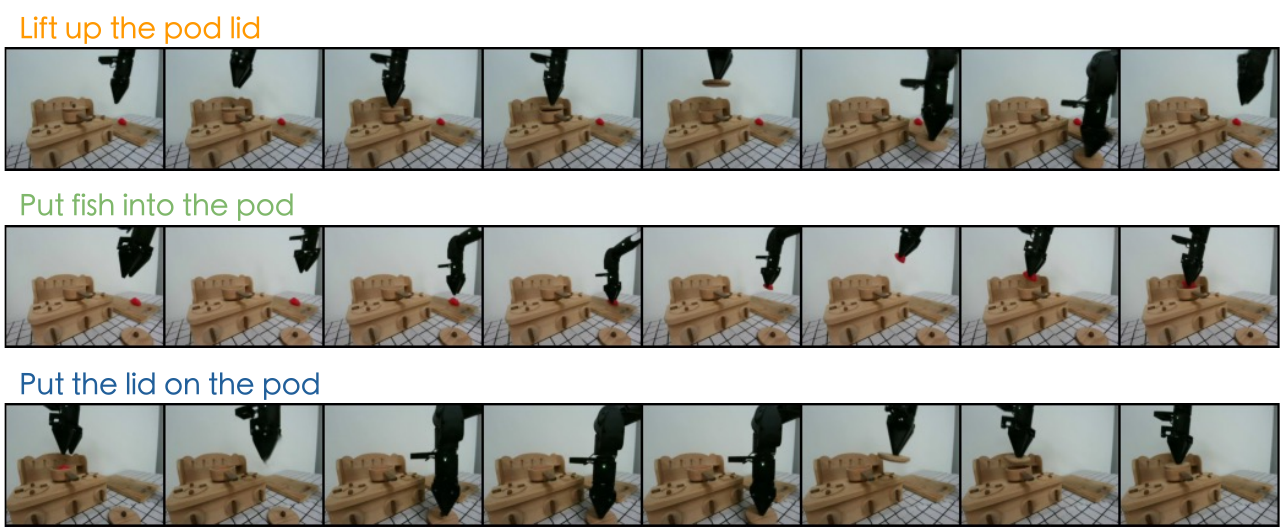}
    \caption{\textbf{Visualization on visual plans in real-world environments.}}
    \label{supp_fig:visualization_real}
\end{figure}

\noindent\textbf{Qualitative analysis.} We also provide qualitative examples of the generated videos paired with corresponding tasks in both CALVIN and real-world environments, in \Cref{supp_fig:visualization_full} and \Cref{supp_fig:visualization_real}, respectively. Our proposed visual planner can reason feasible sub-goals with high temporal coherence.

\begin{figure}[h!]
    \centering
    \includegraphics[width=0.99\linewidth]{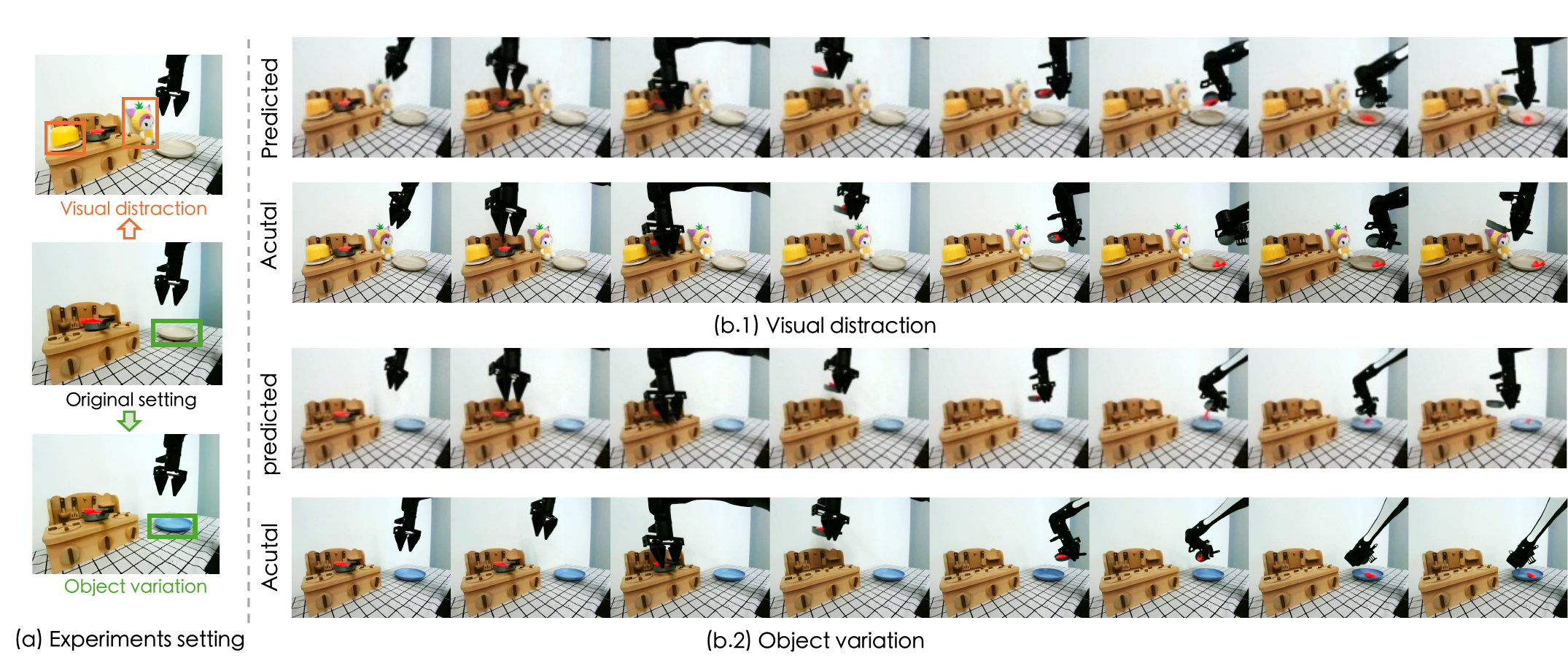}
    \vspace{-3pt}
    \caption{\small \textbf{The predicted visual plan and actual robot execution in real-world experiments.} Our diffusion-based visual planner (up row in each case) can still generate reasonable plans in new scenarios to facilitate successful manipulation (bottom row in each case). 
    }
    \vspace{-4pt}
    \label{fig:generalizability}
\end{figure}

\section{Broader Impact}
\label{appendix-sec:impact}
\vspace{-5pt}

CLOVER contributes to the field of robotics by addressing the limitations of open-loop systems and providing a simple yet robust baseline for closed-loop control. This advancement can inspire further research into adaptive control strategies, error modeling, and real-time feedback mechanisms, pushing the boundaries of robot learning and embodied intelligence.
All our models are trained on publicly available data that is free of private and sensitive information.

\section{License of Assets}
\label{appendix:license}
\vspace{-5pt}

CALVIN~\cite{mees2022calvin} is an open-source simulator which is under the MIT License. The reimplemented methods for performance comparison including ACT~\cite{zhao2023learning}, R3M~\cite{nair2022r3m}, and AVDC~\cite{Ko2023Learning} are all under the MIT License. The pretrained visual encoders adopted, \textit{i.e.}, VC-1~\cite{vc2023}, CLIP~\cite{radford2021clip}, and DINO~\cite{oquab2023dinov2}), are under the CC BY-NC-SA 3.0 US license, MIT license, and Apache License 2.0, respectively.

Our source code and trained models will be publicly available under Apache License 2.0.

\end{document}